\documentclass{article}
\PassOptionsToPackage{numbers, compress}{natbib}

 \usepackage[preprint]{neurips_2026}

\usepackage{bbm}

\usepackage{xcolor}
\definecolor{cvprblue}{rgb}{0.21,0.49,0.74}
\usepackage[pagebackref,breaklinks,colorlinks,citecolor=cvprblue]{hyperref}
\usepackage[accsupp]{axessibility}  

\usepackage[numbers]{natbib}
\usepackage{enumitem}
\usepackage[utf8]{inputenc} 
\usepackage[T1]{fontenc}    
\usepackage{hyperref}       
\usepackage{url}            
\usepackage{booktabs}       
\usepackage{amsfonts}       
\usepackage{nicefrac}       
\usepackage{microtype}      
\usepackage{colortbl}
\usepackage[dvipsnames]{xcolor}
\usepackage{graphicx}
\usepackage{multirow}
\usepackage{caption}
\usepackage{makecell}
\usepackage{arydshln}
\usepackage{array, comment}
\usepackage{nicematrix}
\usepackage{sidecap}
\usepackage{marvosym}

\definecolor{col1}{RGB}{232, 161, 148}
\definecolor{col11}{RGB}{255, 228, 228}
\definecolor{col2}{RGB}{148, 187, 232}
\definecolor{col33}{RGB}{206, 239, 255}
\definecolor{col3}{RGB}{233, 255, 245}
\definecolor{lightgray}{rgb}{0.85,0.85,0.85}
\definecolor{lightlightgray}{rgb}{0.9,0.9,0.9}
\definecolor{verylightBG}{rgb}{0.9,0.99,0.99}
\definecolor{darkgreen}{rgb}{0., 0.7, 0.2}

\usepackage[capitalize]{cleveref}
\crefname{section}{Sec.}{Secs.}
\Crefname{section}{Section}{Sections}
\Crefname{table}{Table}{Tables}
\crefname{table}{Tab.}{Tabs.}

\title{AutoExpert: Automating 3D LiDAR Annotation from Expert-Crafted Guidelines}

\author{%
Yechi Ma$^{1}$,  \quad Wei Hua$^{1}$, \quad Shu Kong$^{2,3,\text{\Letter}}$ 
\\
$^1$Zhejiang University, \quad
$^2$University of Macau, \quad
$^3$IAIBS
\\
\href{https://github.com/cc50121/AutoExpert-3D-benchmark}{project page with code and models}
}

\begin{document}
\maketitle

\begin{abstract}

The contemporary paradigm of scaling data annotation, crucial for developing machine learning solutions, is to hire ordinary human annotators and instruct them with expert-crafted guidelines to label data. 
This paradigm is laborious, tedious, and costly,
motivating us to study an open problem, \emph{auto}-annotation with \emph{expert}-crafted guidelines (dubbed {\em AutoExpert}).
We develop benchmarks by redesigning the evaluation protocol and re-annotating data with nuScenes and PandaSet, 
two 3D detection datasets for autonomous driving research that provide expert-crafted annotation guidelines.
Their guidelines define 18 and 25 object classes, respectively, using nuanced language descriptions and a few visual examples.
Following the guidelines that require using 3D cuboids to label LiDAR data, 
AutoExpert requires algorithms to learn on few-shot labeled images and texts to perform the task of 3D detection on LiDAR data.
Apparently, the challenges of AutoExpert lie in the data-modality and task discrepancy.
Nevertheless, public foundation models (FMs) serve as promising tools to tackle these challenges. 
To address AutoExpert, we adopt a conceptually simple pipeline consisting of three components: (1) 2D object detection and segmentation in RGB images,
(2) lifting 2D detections into 3D using known sensor poses,
and (3) 3D cuboids generation for the 2D detections.
Within this pipeline, we enhance and evaluate a variety of methods such as open-vocabulary detectors, few-shot detectors, and self-supervised learned detectors.
We also develop novel techniques, leading to refined components that boost 3D detection mAP from 12.1 to 25.4 on the AutoExpert-nuScenes benchmark.

\end{abstract}

\section{Introduction}
\label{sec:intro}

Data annotation is a crucial yet costly prerequisite for developing machine learning solutions to numerous applications such as autonomous driving~\cite{caesar2020nuscenes, geiger2013vision, sun2020scalability, Argoverse2}.
The contemporary paradigm of scaling data annotation is through crowd-sourcing \cite{welinder2010multidimensional, zhang2016learning, mac2018teaching}:
hiring ordinary human annotators and instructing them using expert-crafted guidelines to annotate unlabeled data.
Yet, this paradigm is laborious, tedious, and costly.
Moreover,
as ordinary annotators lack domain expertise,
their annotations are often erroneous, subjective, biased, and inconsistent.
Motivated by these, we study an open problem, \textit{auto-annotation with expert-crafted annotation guidelines} (dubbed {\em AutoExpert}).

\begin{figure}[t!]
    \centering
    \includegraphics[width=0.950\linewidth]{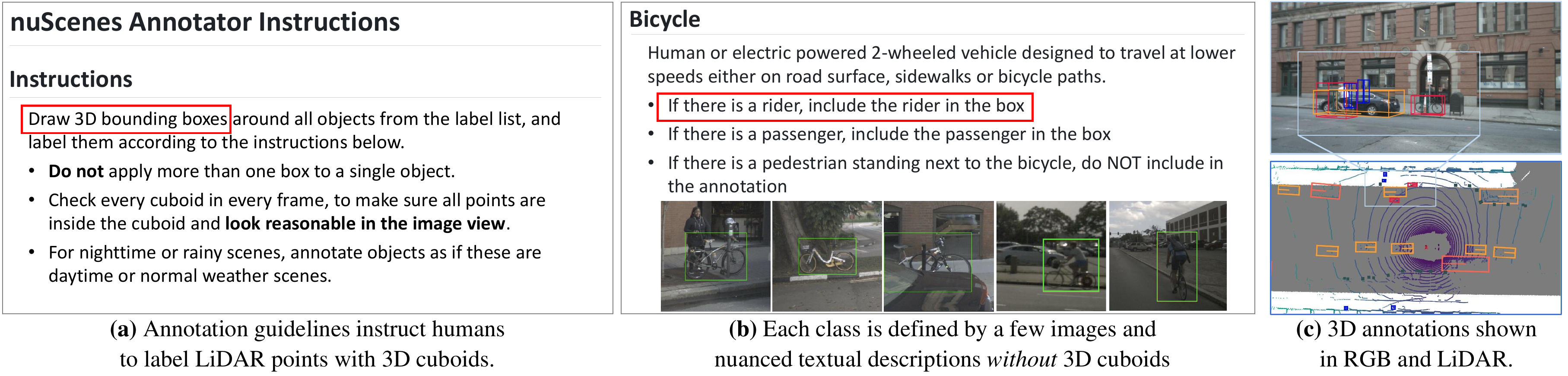}
    \vspace{-2mm}
    \caption{\small
    {\bf Excerpts of the authentic annotation guidelines of the nuScenes dataset}~\citep{caesar2020nuscenes}.
    {\bf (a)} The guidelines instruct human annotators to label LiDAR points with 3D cuboids for specific object classes.
    {\bf (b)} Each class is defined with a few visual examples and nuanced textual descriptions (ref.  \textcolor{Red}{the red box}) \emph{without} 3D annotations.
    Human can comprehend and apply these guidelines to draw 3D boxes.
    {\bf (c)} We visualize the ground-truth or human-annotated 3D cuboids in the RGB image and the Bird's-Eye-View (BEV) of LiDAR points.
    }
\vspace{-4mm}
\label{fig:annotation_guidelines_nuScenes}
\end{figure}

{\bf Status quo.}
While AutoExpert is new, its motivation is shared by several related problems,
such as active learning~\citep{holub2008entropy, settles2009active, kirsch2019batchbald, ren2021survey, bang2024active, wang2025active}, 
open-vocabulary perception~\cite{wu2024towards, zhang2024opensight, etchegaray2024find, zhu2024survey},
and 
few-shot learning (FSL)~\citep{snell2017prototypical, boudiaf2020information, bateni2020improved, satorras2018few, liu2026solving}.
Active learning assumes available ``oracles'' (or domain experts) to label  the data identified as the most informative by the model being trained.
Open-vocabulary perception methods exploit foundation models (FMs),
which have learned common knowledge from web data but do not possess domain expertise and fail to produce results that meet expert-level standards \cite{madan2024revisiting, liu2026visual}.
The recent FSL literature~\citep{madan2024revisiting, liu2025few, liu2026solving} adapts FMs from a perspective of data annotation,
as annotation guidelines generally contain a few visual examples for each defined object class.
Yet, this literature studies FSL methods through over-simplified tasks such as image classification~\citep{liu2025few} and 2D object detection~\citep{madan2024revisiting};
it has not adopted authentic annotation guidelines from real-world applications.

{\bf Benchmarks.}
We explicitly study \emph{AutoExpert} with authentic expert-crafted annotation guidelines.
Specifically, we redesign the benchmarking protocol and re-annotate data with nuScenes~\citep{caesar2020nuscenes} and PandaSet~\cite{xiao2021pandaset}, two established datasets originally designed for 3D detection in autonomous driving research that have released authentic annotation guidelines (\cref{fig:annotation_guidelines_nuScenes}).
The guidelines define 18 and 25 object classes, respectively, using a few visual examples and textual descriptions.
Notably, they do not provide 3D annotations, although they require human annotators to draw 3D cuboids on LiDAR data. Such a discrepancy introduces interesting and distinctive challenges, elaborated next.

{\bf Challenges.}
In our AutoExpert benchmark,
the goal is 3D detection on LiDAR data but the supervision for training comes from a few visual examples and textual descriptions provided in the guidelines, \textit{without} 3D annotations (\cref{fig:annotation_guidelines_nuScenes}).
It is non-trivial to translate the nuanced instructions into action items for machines to follow.
From a research perspective, AutoExpert  might be cast as \emph{multimodal few-shot learning} for 3D detection without 3D annotation.
Moreover,
while it is tempting to leverage FMs, 
there are currently no publicly-available LiDAR-based FMs, making it challenging to apply existing FMs to LiDAR-based 3D detection.
One might adapt multiple appropriate FMs on the visual and textual data for 3D LiDAR detection.

\begin{figure*}[t!]
\centering
\includegraphics[page=1, width=0.95\linewidth]{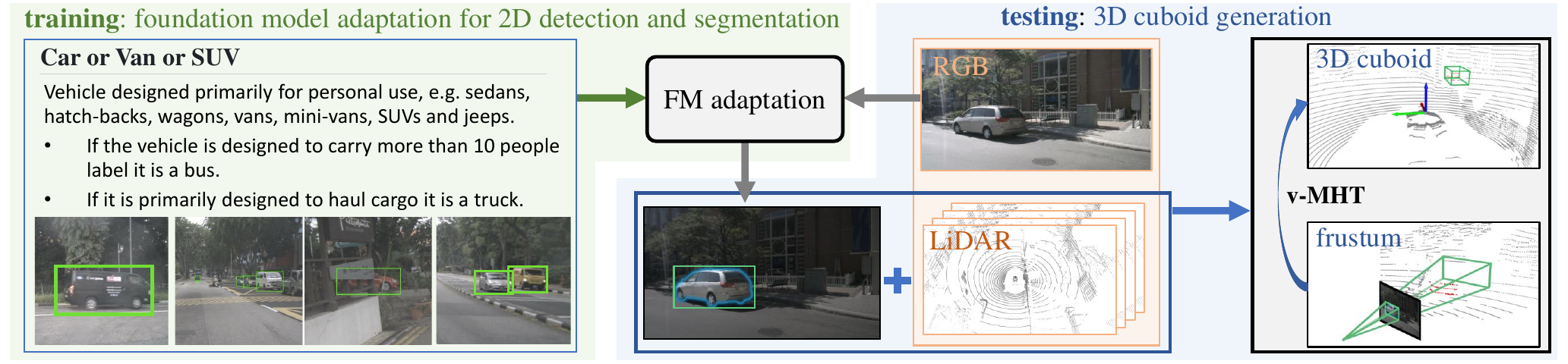}
\vspace{-2mm}
\caption{\small
To solve AutoExpert, we adopt a conceptually simple pipeline. 
Specifically, over the visual examples and textual descriptions that define object classes of interest, we adapt VLMs and VFMs for object detection and segmentation. 
The adapted FMs produce decent 2D detections on unlabeled RGB frames.
With the known parameters of LiDAR and camera,
we lift 2D detections to 3D with our developed VLM-Guided
Multi-Hypothesis Testing (v-MHT) strategy, yielding 3D cuboids as 3D detections.
}
\label{fig:our_pipeline}
\vspace{-4mm}
\end{figure*}

{\bf Methodology.}
To address AutoExpert,
we adopt a conceptually simple pipeline from first principles (\cref{fig:our_pipeline}), consisting of three modules: 
(1) 2D object detection on RGB frames,
(2) LiDAR points association with the 2D detection,
and (3) 3D cuboid generation based on the 2D detection and its associated LiDAR points.
Within this pipeline,
we improve the components by evaluating a variety of methods, such as open-vocabulary detectors, self-supervised detectors, and few-shot learning methods. 
We also develop novel methods to improve 2D detection and 3D cuboid generation.
Our rigorous evaluations and insights lead to multiple techniques that significantly boost 3D detection mAP, e.g., from 6.3 of a baseline to 25.4 on the AutoExpert-nuScenes benchmark (Table~\ref{tab:3d_results_comparison}).

{\bf Contributions.} We make three key contributions:
\begin{enumerate}[leftmargin=15pt, topsep=-2pt, itemsep=3pt, parsep=-2pt]
    \item We introduce a novel and timely task, {\em AutoExpert}, which serve as a facilitator in the development of practical data annotation methods. 
    
    \item We present two AutoExpert benchmarks by re-designing the benchmarking protocol and re-annotating data from the established nuScenes and PandaSet datasets.
    
    \item We address AutoExpert with a conceptually simple pipeline,
    by evaluating a variety of methods, integrating appropriate FMs,
    and developing novel techniques. We significantly boost AutoExpert performance and offer insights to inspire future work.
\end{enumerate}

\section{Related Work}
\label{sec:related-work}

{\bf Data annotation} is a crucial yet costly prerequisite for developing real-world machine learning solutions.
Most works are motivated to reduce annotation cost with human-in-the-loop~\citep{abad2017autonomous, heo2020interactive, cheng2021modular, qiao2023human} or active learning~\citep{holub2008entropy, settles2009active, kirsch2019batchbald, ren2021survey, bang2024active}.
But they over-simplify the complexity of data annotation by exclusively focusing on common objects (e.g., {\tt car} and {\tt person})~\citep{ramanan2003automatic, reza2025segbuilder, wu2024efficient, wang2006annosearch, zhou2024openannotate3d, ma2026roadside}.
They also neglect the fact that real-world applications annotate data in a safety-critical way \citep{madan2024revisiting, chakravarthy2025lidar}.
For example,
{\tt bicycle} is defined differently~\citep{caesar2020nuscenes} from what one would have thought (\cref{fig:annotation_guidelines_nuScenes}): it includes the rider if existing, rather than solely the bicycle itself.
Hence, it remains an open question how to automate annotation of domain-specific data directly from expert-crafted annotation guidelines,  namely AutoExpert.
Our work explores AutoExpert.

{\bf Foundation Models (FMs)} are key to modern AI agents such as GPT \citep{achiam2023gpt}, Gemini \citep{team2023gemini} and Qwen \citep{qwen}.
We briefly review Vision-Language Models (VLMs) and Visual Foundation Models (VFMs), which are used in this paper.
VLMs are pretrained on large-scale image-text pairs~\citep{radford2021learning, jia2021scaling, liu2023llava, liu2023grounding, qwen}, excelling at visual understanding tasks such as visual grounding, image captioning, and visual question answering.
VFMs are trained on visual data~\citep{chen2020simple, caron2021emerging, zhou2021ibot, oquab2023dinov2, touvron2022deit} and excel at perception tasks such as object detection~\citep{zhang2022dino, liu2023grounding, ren2024grounded, liu2023grounding} and segmentation~\citep{kirillov2023segment, wang2024segment}.
Recent works exploit VLMs to LiDAR perception by lifting VLM's 2D detections on images to 3D LiDAR points using known camera-LiDAR sensor parameters~\citep{khurana2024shelf, ovsep2024better, zhang2024opensight, etchegaray2024find}. 
Yet, they fall short in domains like autonomous driving, 
where object definitions require nuanced understanding critical to safety~\citep{madan2024revisiting}.
Therefore, adapting FMs effectively to address 3D detection for autonomous driving remains an open problem.
Our work contributes to this area by exploring AutoExpert in the autonomous driving scenario.

{\bf 3D LiDAR detection} is a crucial task in autonomous driving, establishing multiple datasets~\cite{geiger2013vision, caesar2020nuscenes, sun2020scalability, xiao2021pandaset, wilson2023argoverse}.
Among them, 
only nuScenes and PandaSet have released their \emph{official annotation guidelines}; others released instructions to familiarize users with their challenges. 
Most methods of 3D LiDAR detection train detectors over massive 3D annotated LiDAR data,
optionally exploiting annotated RGB frames~\citep{yin2021center, bai2022transfusion, li2024bevformer, chow2025ov}.
Some study training detectors without 3D annotations,
in an unsupervised or self-supervised manner~\citep{zhang2023towards, wu2024commonsense}.
Recent methods explore open-vocabulary 3D detection \cite{zhang2024opensight, etchegaray2024find} by exploiting FMs without 3D annotated data.
\cite{madan2024revisiting, peri2023towards} note two issues.
First,
FMs fail to produce results that meet expert-level or safety-critical standards,
due to a discrepancy between task-specific expertise and common knowledge learned by FMs from web data \cite{madan2024revisiting}.
Second, 
existing methods exclusively focus on common objects (e.g., {\tt car} and {\tt cyclist}) and neglect rare but safety-critical ones (e.g., {\tt stroller} and {\tt wheelchair}),
whereas annotation guidelines have defined all these classes~\citep{peri2023towards}.
Our AutoExpert considers all the classes for benchmarking.

{\bf Few-Shot Learning} (FSL) aims to develop methods to learn from a small number of labeled examples~\citep{snell2017prototypical, boudiaf2020information, bateni2020improved, satorras2018few}.
Recent FSL methods adapt VLMs~\citep{tipadapter, zhang2023prompt, lin2023multimodality, tang2024amu, clap24, maple, liu2025few}.
Some point out that FSL is better studied from a data annotation perspective, 
as annotation guidelines realistically provide few-shot visual examples for demonstration~\citep{madan2024revisiting, liu2025few}.
However, 
the current FSL literature has largely focused on oversimplified tasks such as image classification~\citep{madan2024revisiting} and 2D object detection~\citep{liu2025few}, and exploited a single FM.
In contrast, AutoExpert is a more challenging setting that requires developing multimodal few-shot learning methods for 3D LiDAR detection without 3D annotation.
The potential solutions are expected to ensemble multiple appropriate FMs.
Importantly, AutoExpert holds practical significance as its evaluation protocol is grounded in real-world annotation practices, making use of authentic annotation guidelines.

\section{AutoExpert: Problem Formulation and Benchmarking Protocol}
\label{sec:problem-protocol}

{\bf Problem formulation.} 
AutoExpert emphasizes human's capability of producing 3D cuboids by learning from expert-crafted guidelines, which provide descriptions and a few images to define each object class (\cref{fig:annotation_guidelines_nuScenes}).
Formulating human annotators' workflow,
potential methods should
(1) comprehend the descriptions and visual examples to understand each object class,
(2) detect objects in RGB frames and associate LiDAR points with them, 
(3) utilize prior knowledge about objects' 3D shapes and sizes to generate proper 3D cuboids in the LiDAR data.
Hence, AutoExpert evaluates methods w.r.t 3D detection metrics, complemented with 2D detection metrics:
\begin{itemize}[leftmargin=15pt, topsep=-2pt, itemsep=3pt, parsep=-2pt]
\item     
    {\em 3D metrics.}
    Following \citep{caesar2020nuscenes}, we report mean Average Precision  over per-class AP at ground-plane distance thresholds: [0.5, 1.0, 2.0, 4.0] in meters.
    We denote this metric as mAP$^{3D}$.
    We also report the NDS metric, which summarizes mAP$^{3D}$, the average of per-class translation error (ATE), scale error (ASE), orientation error (AOE), velocity error (AVE), and attribute error (AAE).
    
\item  
    \emph{2D metrics.}
    Following \citep{lin2014coco}, we report mean Average Precision that averages per-class AP at IoU=0.5.
    We denote this metric as mAP$^{2D}$.
\end{itemize}

\begin{figure}[t!]
\centering
\includegraphics[page=1, width=1.0\linewidth]{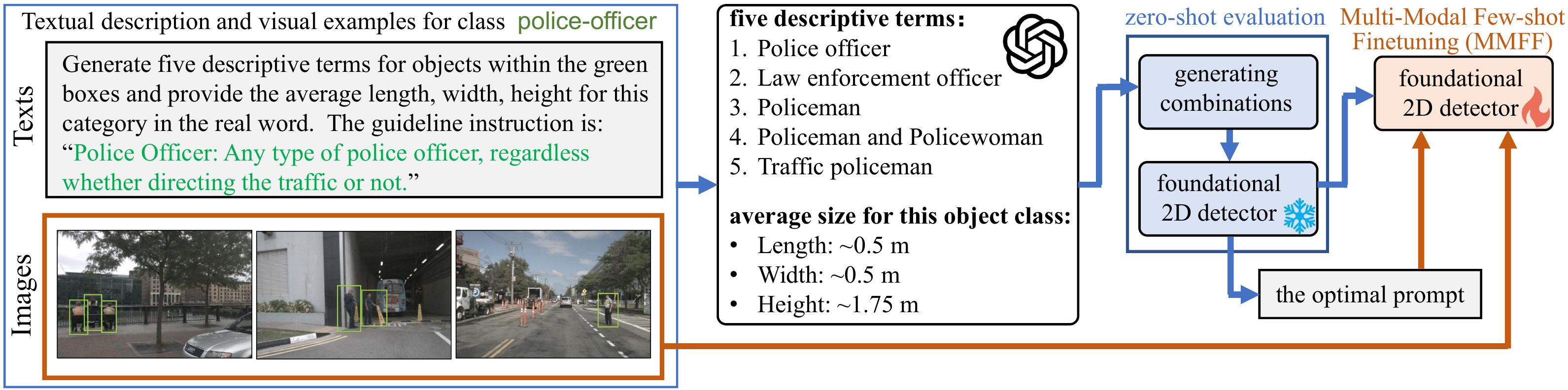}
\vspace{-5mm}
\caption{\small
For each class name, we use a VLM (e.g., GPT-4o~\citep{achiam2023gpt} and Qwen~\cite{qwen}) to find a list of terms that match its description and visual examples in the annotation guidelines. 
We select the term or combined term that yields the best zero-shot detection performance of a foundational detector (e.g., GroundingDINO~\citep{liu2023grounding}) on the valset.
We construct a multimodal few-shot train-set using the selected terms and the images provided in annotation guidelines to finetune the detector, yielding notable improvements (\Cref{tab:2D_comparison}).
}
\vspace{-3mm}
\label{fig:Prompt_refining}
\end{figure}

{\bf Benchmark Datasets.}
Based on two established datasets, nuScenes \cite{caesar2020nuscenes} and PandaSet \cite{xiao2021pandaset},
we construct two benchmarks named \textit{AutoExpert-nuScenes} and \textit{AutoExpert-PandaSet}.
Both datasets were designed for 3D detection research in autonomous driving and licensed under CC BY-NC-SA 4.0,
and have released their annotation guidelines that define 18 and 25 object classes, respectively.
While the guidelines contain visual images to demonstrate each class (\cref{fig:annotation_guidelines_nuScenes}),
we do not use them in our benchmarks as they were likely sourced from the Internet that potentially have copyright issues.
Therefore,
from their official train-sets, 
we select 4-8 images, which capture clear visual signatures of objects to simulate iconic examples displayed in annotation guidelines \cite{madan2024revisiting}, as those provided by annotation guidelines.
The selected images can serve as a federated train-set,
where each image is annotated exclusively for a single target class \cite{madan2024revisiting}.
For example, 
the selected images for {\tt police}-{\tt officer} (\cref{fig:Prompt_refining}) contain {\tt car} and {\tt truck} which are not annotated.
Furthermore, we re-annotated PandaSet as we observe notable misalignment of 3D annotations and RGB frames, e.g., many ghosting 3D cuboids do not correspond to any objects in RGB frames (ref. Appx.~\S\ref{sec:pandaset-appendix}).
Our \href{https://github.com/cc50121/AutoExpert-3D-benchmark}{devkit} and \href{https://huggingface.co/spaces/autoexpert-cvpr2026-workshop/auto3D}{annotations} are released under the MIT License.

{\bf Train, validation, and test sets}.
In our benchmarks,
we consider the images and texts in the annotation guidelines as the train-sets.
In AutoExpert-nuScenes,
we sample 570 frames from nuScenes official train-set as the val-set
and use its official validation set (6,019 frames) as the test-set.
In AutoExpert-PandaSet,
we use 8 and 192 of our annotated RGB frames (and the corresponding LiDAR sweeps) as the val-set and test-set, respectively.
In the benchmarks, the val-sets simulate the real-world practice for expert-in-the-loop quality control,
as domain experts typically oversee annotation progress.
We focus on AutoExpert-nuScenes in the main paper and provide details about AutoExpert-PandaSet in Appx.~\S\ref{sec:pandaset-appendix} for two reasons: (1) we use the latter for challenge competition where we held out the test-set, and (2) conclusions are consistent on both benchmarks.

\section{Methodology: Baselines and Improvements}
\label{sec:method}

We explore AutoExpert methods from first principles with a conceptually intuitive pipeline (\cref{fig:our_pipeline}), consisting of three modules:
(1) 2D object detection on RGB frames,
(2) LiDAR points association with the 2D detection,
and (3) 3D cuboid generation based on the 2D detection and its associated LiDAR points.
Within this pipeline, many existing 2D detectors and 3D proposal detectors can serve a module.
Hence, baseline methods can be combinations of strong foundational 2D and 3D detectors \cite{peri2023towards, ma2025long},
or open-vocabulary 3D detectors \cite{lu2023open, zhang2024opensight, etchegaray2024find, sun2025annofreeod, lee2025openbox} (which also exploit such a pipeline).
Notably, the open-vocabulary detectors rely on off-the-shelf FMs to produce class labels, which do not meet experts' annotations \cite{madan2024revisiting} (\Cref{tab:3d_results_comparison}), as FMs do not possess domain expertise.
Therefore,
in this section,
we develop novel and improved methods to improve 2D detection in \S\ref{ssec:2D-detection} and 3D detection in \S\ref{sec:3D-cuboid-search},
as well as additional techniques for further improvements \S\ref{ssec:tech-endeavor}.

\subsection{Improving 2D Detection by Multimodal Few-Shot Finetuning}
\label{ssec:2D-detection}
For 2D detection on RGB frames, we start with the foundational detector GroundingDINO~\citep{liu2023grounding},
which yields impressive zero-shot detection performance on common objects but is not tailored to specific tasks~\citep{madan2024revisiting}.
For example, for autonomous driving as in nuScenes,
{\tt bicycle} is defined differently from common sense (\cref{fig:annotation_guidelines_nuScenes}): annotators should include the existing rider in the box annotation in light of  driving safety.
Hence, we adapt GroundingDINO to AutoExpert with techniques below.

\textbf{Prompt engineering} often require manual designs of prompts to enhance zero-shot performance.
In this work, we present an \emph{automated} method to generate better prompts (\cref{fig:Prompt_refining}) to enhance 3D detection.
Specifically, 
for an object class, 
we prompt a VLM (e.g., GPT and Qwen)
to find five descriptive terms or synonyms \cite{parashar2023prompting} that fit its corresponding description and images provided in the guidelines.
Then, we use the terms and their combinations to prompt GroundingDINO for zero-shot detection.
We pick the term or a combination that yields the highest 2D detection precision of this class on the valset.
As shown in \Cref{tab:2D_comparison},
using the selected terms as the refined class names in prompts
remarkably improves detection performance.

\begin{figure}[t]
  \centering
  \begin{minipage}[c]{0.63\linewidth}
    \centering
\includegraphics[width=0.99950\linewidth, clip=true, trim = 0mm 0mm 95mm 0mm]{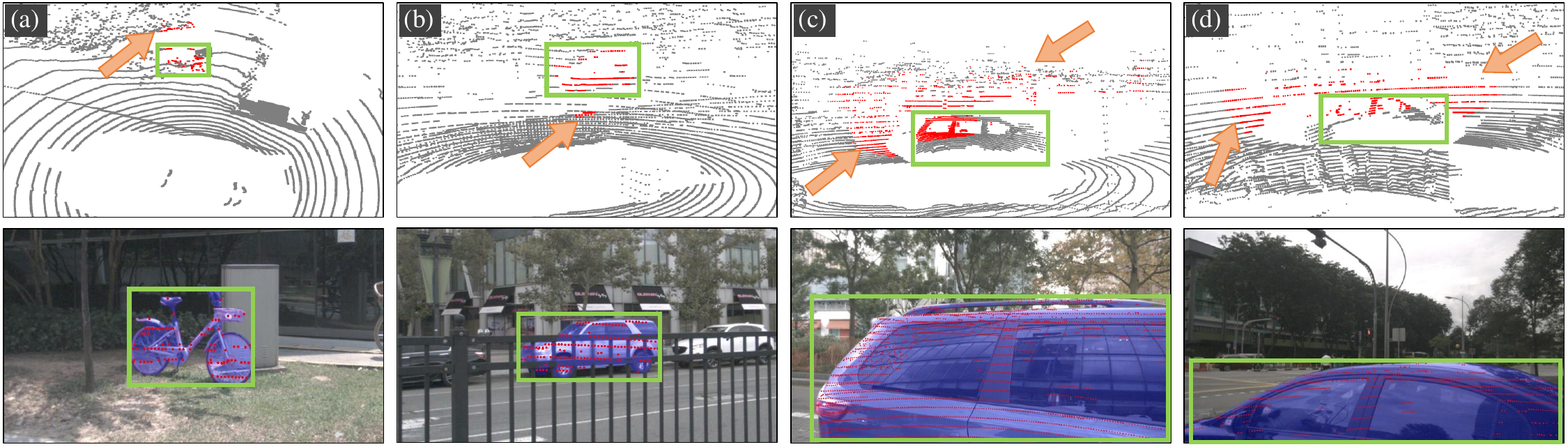}
  \end{minipage}
  \hfill
  \begin{minipage}[c]{0.36\linewidth}
    \small
    \vspace{3mm}
    \caption{\small
    Generating 3D cuboids based on LiDAR points is challenging as points can be from occluders and backgrounds.
    For example, (a) LiDAR points projected on a {\tt bicycle} foreground mask can be from the background scene through wheels;
    (b) points projected on a {\tt car} mask can be from an occluding fence;
    (c) points projected on a {\tt car} mask can be background through the windows.
    }
    \label{fig:challenge_3D_grouping}
  \end{minipage}
  \vspace{-5mm}
\end{figure}

\textbf{Multimodal few-shot finetuning.}
As a small amount of images are available in the annotation guidelines, 
we use them, along with selected terms in the last step,
as few-shot training data to finetune the foundational detector (\cref{fig:Prompt_refining}).
It is worth noting that visual examples in the guidelines are annotated in a federated way, 
i.e., 
all objects belonging to the focused class are annotated while others are not.
Therefore,
when finetuning on the few-shot images,
we compute the loss pertaining to the specific classes without penalizing detections for other classes as false positives.
As shown in \Cref{tab:2D_comparison},
such multimodal few-shot finetuning notably improves 2D and 3D detection.

\subsection{Improving 3D Detection via VLM-Guided Multi-Hypothesis Testing}
\label{sec:3D-cuboid-search}
To produce a 3D cuboid for a 2D detection,
prior methods \cite{wu2024efficient, khurana2024shelf, zhou2024openannotate3d, lee2025openbox, sun2025annofreeod} apply a 3D proposal detector or HDBSCAN for clustering points within the frustum defined by this 2D detection (ref. bottom-right of \cref{fig:our_pipeline}), optionally after running SAM~\cite{kirillov2023segment} to obtain foreground segmentation and remove noisy LiDAR points. 
However,
the remaining LiDAR points can come from background or occluders (\cref{fig:challenge_3D_grouping}), making it non-trivial to generate a 3D cuboid with appropriate dimension to fit this detected object.
In this section,
we present a strategy to mitigate this issue.
Our idea is to first construct a per-instance cuboid and then search for a placement in 3D.
The dimension of the cuboid is fixed during searching,
helping mitigate the influence caused by noisy LiDAR points from occluders and background.
Concretely, our strategy first leverages a VLM as a \textit{virtual expert} to infer dimensions and orientation of the detected object in 2D frame,
then constructs a cuboid based on them,
and lastly performs Multi-Hypothesis Testing (MHT) to search for a 3D placement towards the final 3D cuboid.
We name the strategy (virtual expert) VLM-Guided MHT, dubbed \textbf{v-MHT}.

\begin{figure*}[t!]
\centering
\includegraphics[width=.950\linewidth]{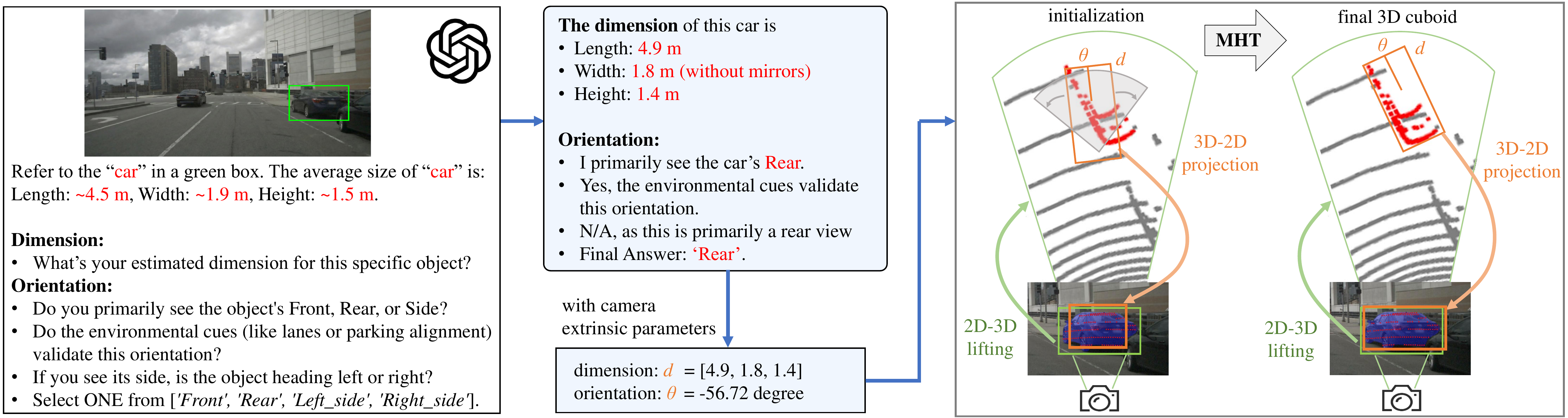}
\vspace{-1mm}
\caption{\small
\textbf{Overview of the v-MHT strategy for 3D cuboid generation.} 
It begins by prompting a VLM to infer the 3D information about a target 2D detection (cf. the left panel), including its 3D dimension $d$ and orientation.
As we find it challenging to directly prompt the VLM to output an orientation angle,
we instruct the VLM to output the location of the object in the image and the visible parts of this object,
With the output and the known camera extrinsic parameters, we estimate the object orientation $\theta$ (cf. the mid panel).
Lastly,
we initialize a 3D cuboid with the estimated dimension $d$ and orientation $\theta$ and perform multi-hypothesis testing (MHT) to search for the final cuboid that best fits LiDAR points and the 2D detection box (cf. the right panel).
}
\vspace{-2mm}
\label{fig:v-MHT-box-fitting}
\end{figure*}

\textbf{Virtual expert reasoning.}
We employ a VLM (e.g., GPT and Qwen) to reason about the dimension $d$ and orientation $\theta$ for \textit{individual} detected objects in RGB image.
Specifically, we first compose a prompt containing
(1) the RGB image with the detected object highlighted by a green bounding box, 
(2) a textual description about the class-specific average size  (obtained in \cref{fig:Prompt_refining}), 
and (3) a question to estimate the dimension and orientation of the target object.
We use the prompt to instruct VLM to estimate 3D shape/dimension $d$ and orientation $\theta$ for this object. 
Refer to \cref{fig:v-MHT-box-fitting} for an illustration of this step and
Appx.~\S\ref{sec:prompt-VL-MHT-appendix} for a full prompt.
We reiterate that this step produces per-instance cuboids that facilitate 3D cuboid generation for individual 2D detections (ref. Appx.~\Cref{tab:MHT-v-MHT} for an analysis),
e.g., better fitting a sedan and a van, both of which belong to the class ``car'' but have distinct sizes.

\textbf{MHT refinement.}
We initialize a cuboid defined by the dimension $d$ and orientation $\theta$ estimated in the previous step.
Then, 
in the frustum determined by the detection and camera parameters,
we perform MHT to search for a placement of this cuboid,
by selecting a hypothesis that maximizes joint objectives (\cref{fig:v-MHT-box-fitting}-right):
(1) coverage of foreground LiDAR points,
and (2) Intersection-over-Union (IoU) between the projected hypothetic 3D cuboid and the 2D detection box.
The search space is constrained in translation and rotation,
around the estimated location and orientation from the previous step.
This constrained search space not only improves searching efficiency,
but also resolves the $180^\circ$ ambiguity, avoiding converging to flipped orientations caused by the geometric symmetry of an object.
It is worth noting that our implementation is highly efficient by utilizing the Numba compiler~\citep{lam2015numba} and GPU parallelization.
Refer to  Appx.~\S\ref{sec:details-3d-generator-appendix} and \href{https://github.com/cc50121/AutoExpert-3D-benchmark/blob/main/src/masks_to_3D_results.py}{our code} for details, and Appx.~\S\ref{sec:comp_efficiency} for computation analysis.

\subsection{Techniques for Further Improvements}
\label{ssec:tech-endeavor}

We present several simple and effective techniques to further improve 3D detection, elaborated below.

\textbf{LiDAR sweep aggregation.}
A single LiDAR sweep can be too sparse to precisely localize objects in 3D.
Existing 3D detectors aggregate \emph{history} sweeps at a given timestamp;
AutoExpert allows aggregating ``\emph{future}'' sweeps in light of offline data annotation.
We analyze per-class 3D detection performance with different aggregation strategies (\Cref{tab:sweep_aggregation}), and find that specific strategies can notably improve 3D detection on certain classes.
We also tested per-instance aggregation \cite{baur2024liso, wu2024commonsense, zhang2023towards}, 
which aggregates multiple sweeps if the detected object is static determined by its 3D tracking trajectory (detailed later).
However,
this aggregation did not rival per-class aggregation,
primarily owing to the robustness of our v-MHT strategy, 
which uses fixed instance-specific cuboids (to search for a 3D placement as the generated 3D cuboid) that effectively mitigate the effect caused by smears in aggregated sweeps (ref. detailed study in Appx.~\S\ref{sec:sweep_aggregation}).

\textbf{3D cuboid scoring with geometric cues.}
In addition to use 2D detection confidence $S_{\text{2D}}$ 
for scoring a generated 3D cuboid,
we further incorporate 3D geometric cues,
measured by occupancy rate~\citep{wu2024commonsense} based on LiDAR point distribution within the cuboid.
Specifically,
we (1) project the 3D cuboid onto the ground plane, obtaining a BEV rectangle;
(2) discretize this rectangle into a 7$\times$7 grid;
(3) count grid cells ($N$) that contain at least one project LiDAR point;
and
(4) compute the occupancy rate as $S_{\text{3D}} = N/7^{2}$.
We define the final score  
$S = \alpha* S_{\text{2D}} + (1-\alpha) * S_{\text{3D}}$,
where the hyperparameter $\alpha$ is set w.r.t the highest 3D detection mAP on the val-set.
\Cref{tab:3D_ablation} shows the notable benefit of this score.

\textbf{Tracking-based score refinement.}
For a generated 3D cuboid at a given timestamp,
we leverage temporal information to refine its score.
Over consecutive LiDAR sweeps, 
we heuristically associate spatially-close 3D cuboids that share the same predicted class.
With the track, we replace scores of individual 3D cuboids with their mean.
Despite its simplicity, this method remarkably improves 3D detection performance (\Cref{tab:3D_ablation}).
We also examined an other method that uses SAM2 \cite{ravi2024sam} for 2D tracking and averages detection scores within the track;
but it was not competitive (cf. Appx.~\S\ref{sec:comparison-3D-2D-tracking-appendix}).

\begin{figure*}[t]
\centering
\includegraphics[width=0.950\linewidth]{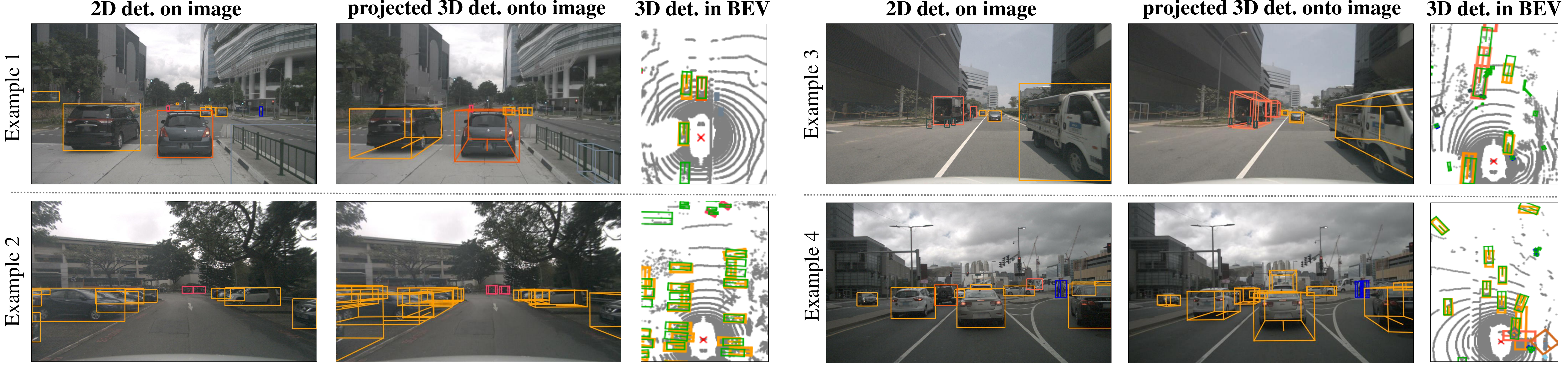}
\vspace{-1mm}
\caption{\small
\textbf{Visual results of auto3D on four testing examples of AutoExpert-nuScenes.}
For each example, we display 2D and 3D detections (i.e., the generated 3D cuboids) on the RGB image and the BEV of LiDAR data.
Results show that auto3D decently detects objects that are in far field and small in size, which are challenging to detect~\citep{peri2023towards, gupta2023far3det}.
Refer to Appx.~\S\ref{sec:more-visualizations-appendix} for more visual results and the \href{https://github.com/cc50121/AutoExpert-3D-benchmark/blob/main/demo-annoexpert-infer-vis.ipynb}{Jupyter Notebook} for a demo.
}
\vspace{-0mm}
\label{fig:visual-results}
\end{figure*}

\begin{table}[t]
\centering
\small
\begin{minipage}[c]{0.61\linewidth}
\setlength{\tabcolsep}{0.9mm}
\scalebox{0.80}{
\begin{tabular}{llccccccc}
\toprule
Method & pub. & \cellcolor{col11}mAP$^{3D}$ & \cellcolor{col33}NDS & ATE  & ASE & AOE  & AVE & AAE    \\
\midrule

Oyster~\citep{zhang2023towards} & \scriptsize\color{gray}CVPR'23 & \ \ \cellcolor{col11}6.3 & \cellcolor{col33}10.7 & 0.755 & 0.715 & 1.451 & 1.201 & 0.771 \\
\quad w/ frustum & ours & \cellcolor{col11}12.9 & \cellcolor{col33}15.6 & 0.723 & 0.651 & 1.563 & 1.008 & 0.711  \\

LISO~\citep{baur2024liso}  & \scriptsize\color{gray}ECCV'24 & \ \ \cellcolor{col11}8.9 & \cellcolor{col33}13.1 & 0.725 & 0.679 & 1.411 & 1.152 & 0.733  \\
\quad  w/ frustum   & ours & \cellcolor{col11}15.7 & \cellcolor{col33}19.8 & 0.679 & 0.583 & 1.429 & 0.903 & 0.644  \\

UNION~\citep{lentsch2024union}  & \scriptsize\color{gray}NeurIPS'24 & \cellcolor{col11}9.7 & \cellcolor{col33}13.8 & 0.726 & 0.667 & 1.412 &  1.112 & 0.713  \\

\quad  w/ frustum  & ours & \cellcolor{col11}16.8 & \cellcolor{col33}20.6 & 0.677 & 0.579 & 1.397 & 0.901 & 0.622  \\

CPD~\citep{wu2024commonsense}    & \scriptsize\color{gray}CVPR'24 & \cellcolor{col11}10.1 & \cellcolor{col33}14.2 & 0.718 & 0.658 & 1.408 &  1.108 & 0.708  \\
\quad  w/ frustum  & ours & \cellcolor{col11}17.9 & \cellcolor{col33}22.3 & 0.641 & 0.549 & 1.338 & 0.886 & 0.590  \\
\midrule
Find\&Prop~\citep{etchegaray2024find} & \scriptsize\color{gray}ECCV'24 & \cellcolor{col11}11.3 & \cellcolor{col33}16.0 & 0.756 & 0.542 & 1.191 & 1.091 & 0.671  \\

\quad  w/ ft-GD  & ours & \cellcolor{col11}17.1 & \cellcolor{col33}21.1 & 0.660 & 0.540 & 1.229 & 0.901 & 0.645  \\
OpenSight~\citep{zhang2024opensight} & \scriptsize\color{gray}ECCV'24 & \cellcolor{col11}11.8 & \cellcolor{col33}16.4 & 0.752 & 0.539 & 1.179 & 1.088 & 0.655  \\

\quad  w/ ft-GD  & ours & \cellcolor{col11}14.6 & \cellcolor{col33}18.2 & 0.733 & \textbf{0.530} & 1.109 & 1.001 & 0.645  \\

OpenBox \citep{lee2025openbox} & \scriptsize\color{gray}NeurIPS'25 & \cellcolor{col11}10.6 & \cellcolor{col33}15.5 & 0.778 & 0.612 & 1.225 & 1.092 & 0.591  \\
\quad  w/ ft-GD   & ours & \cellcolor{col11}15.2 & \cellcolor{col33}20.5 & 0.651 & 0.580 & 1.201 & 0.924 & 0.560  \\

AnnofreeOD \citep{sun2025annofreeod} & \scriptsize\color{gray}ICCV'25 & \cellcolor{col11}11.7 & \cellcolor{col33}16.2 & 0.781 & 0.589 & 1.193 & 1.086 & 0.597  \\
\quad  w/ ft-GD   & ours & \cellcolor{col11}17.3 & \cellcolor{col33}22.1 & 0.648 & 0.566 & 1.181 & 0.899 & 0.543  \\

CM3D \citep{khurana2024shelf} & \scriptsize\color{gray}CoRL'24 & \cellcolor{col11}12.1 & \cellcolor{col33}16.6 & 0.775 & 0.587 & 1.189 & 1.084 & 0.579  \\
\quad  w/ ft-GD   & ours & \cellcolor{col11}18.2 & \cellcolor{col33}23.1 & 0.636 & 0.543 & 1.322 & \textbf{0.875} & 0.548  \\

\midrule

\textbf{auto3D} & ours & \cellcolor{col11}\textbf{25.4} & \cellcolor{col33}\textbf{27.2} & \textbf{0.552} &  0.534 & \textbf{0.992} & 0.927 & \textbf{0.536}  \\
\bottomrule
\end{tabular}
}
\end{minipage}
\hfill
\begin{minipage}[c]{0.38\linewidth}
\vspace{-0mm}
\caption{\small
{\bf Benchmarking on AutoExpert-nuScenes}.
The upper panel lists methods based on self-supervised 3D proposal detectors, which do not output class labels.
We assign them with the class labeled predicted by our finetuned GroundingDINO (with refined class names).
Moreover,
we also empower them by providing our generated frustum (marked by ``w/ frustum''),
helping them achieve significant boosts.
The mid panel lists methods based on open-vocabulary 3D detectors.
We empower them by replacing their 2D detectors with our finetuned GroundingDINO with refined names (marked by ``w/ ft-GD'').
Nevertheless, our method auto3D significantly outperforms all the compared methods.
}
\label{tab:3d_results_comparison}
\end{minipage}
\vspace{-3mm}
\end{table}

\begin{table}[t]
\centering
\begin{minipage}[c]{0.40\linewidth}
\setlength{\tabcolsep}{1.2mm}
\scalebox{0.85}{
\begin{tabular}{r c ccc}
\toprule
 & name & mAP$^{2D}$ & mAP$^{3D}$ & NDS \\
\midrule
\rowcolor{lightgray}Detic & o-name & 16.5 & 12.1 & 16.6 \\
GD & o-name & 16.9 & 16.1 & 21.3 \\
ft-GD & o-name & 20.0 & 16.6 & 21.2 \\
GD & r-name & 18.2 & 15.7 & 22.1 \\
ft-GD & r-name & \textbf{20.8} & \textbf{18.2} & \textbf{23.1} \\
\bottomrule
\end{tabular}
}
\end{minipage}
\hfill
\begin{minipage}[c]{0.58\linewidth}
\vspace{-2mm}
\caption{\small
{\bf Comparison of 2D detectors} w.r.t 2D and 3D metrics.
Detic (prompted with original name, o-name) \cite{zhou2022detecting} is the 2D detector used in CM3D \cite{khurana2024shelf}, 
serving as the baseline in this table.
We test the performance of prompting or finetuning GroundingDINO (GD) with original class names vs. refined names (r-name) given by the GPT-4o model (\cref{fig:Prompt_refining}).
Results show that finetuning GD (ft-GD) using r-name performs the best.
}
\label{tab:2D_comparison}
\end{minipage}
\vspace{-4mm}
\end{table}

\section{Experimental Results and Analysis} 
\label{sec:exp}

We term our final method as \textbf{auto3D} by incorporating all the developed techniques in \S\ref{sec:method}.
Hereafter, 
we conduct extensive experiments on the AutoExpert-nuScenes benchmark to validate auto3D and ablate its core components; 
we report the results on AutoExpert-PandaSet in Appx.~\S\ref{sec:pandaset-appendix} where the same conclusions hold.
\cref{fig:visual-results} visually displays the 3D detections of auto3D. 
Again, we note that baseline methods for AutoExpert can be modified open-vocabulary detectors and self-supervised 3D detectors, once being enhanced with appropriate 3D proposal detectors and 2D object detectors, respectively.
We begin this section by introducing the compared / baseline methods.

\textbf{Compared methods.}
We explain how we enhance existing methods of self-supervised 3D proposal detectors and open-vocabulary 3D detectors to solve AutoExpert, following the pipeline in \cref{fig:our_pipeline}:
\begin{itemize}[leftmargin=15pt, topsep=-2pt, itemsep=3pt, parsep=-2pt]
\item 
    \textit{Self-supervised 3D proposal detectors}
    UNION \citep{lentsch2024union},
    Oyster~\citep{zhang2023towards}, CPD~\citep{wu2024commonsense}, and 
    LISO~\citep{baur2024liso} share a common philosophy that clusters LiDAR points using DBSCAN,
    fits each cluster with a 3D cuboid \citep{zhang2017efficient, you2022learning},
    and self-trains a network towards a 3D proposal detector.
    Note that they do not produce class labels as they work only on unlabeled LiDAR data.
    To solve AutoExpert, 
    we assign their 3D proposals with class labels predicted by the finetuned GroundingDINO with refined names (dubbed ft-GD hereafter; cf. \Cref{tab:2D_comparison}):
    projecting each 3D proposal onto the image plane,
    finding a matched ft-GD's 2D detection,
    and assigning this 2D detection's predicted class label to the 3D proposal.
    Furthermore, as our auto3D exploits frustums,
    we also improve these approaches by leveraging frustums:
    based on each ft-GD's 2D detection,
    we define a frustum in LiDAR point cloud and run these methods within this frustum to produce a 3D proposal,
    which naturally inherits the predicted class label of the 2D detection.
    In other words, we use frustums to provide more targeted searching space and help these methods perform better (as shown in \Cref{tab:3d_results_comparison}).
    To distinguish their variants,
    we append ``w/ frustum'' to these methods when reporting results.
    Appx.~\S\ref{sec:more-details-3d-appendix} describes more detailed implementation.

\item 
    \textit{Open-vocabulary 3D detectors}
    CM3D \cite{khurana2024shelf},
    OpenSight \cite{zhang2024opensight},
    Find\&Propagate \cite{etchegaray2024find},
    OpenBox \cite{lee2025openbox} 
    and AnnofreeOD \cite{sun2025annofreeod}
    train 3D detectors by exploiting both RGB and LiDAR data.
    They make use of off-the-shelf 2D object detectors such as
    Detic \cite{zhou2022detecting},
    GroundingDINO \cite{liu2023grounding} and OWL-ViT \cite{minderer2022simple} to produce 2D detections with predicted class labels.
    We improve them by replacing their off-the-shelf 2D detectors with our finetuned GroundingDINO (with refined names).
     These methods adopt the pipeline in \cref{fig:our_pipeline}, commonly rely on and DBSCAN to bottom-up cluster points and per-class cuboid priors to generate 3D cuboids.
    Specifically, 
    CM3D 
    relies on HD maps and lanes to produce 3D cuboids;
    OpenSight stores a bank of 3D priors to facilitate 3D cuboid generation;
    Find\&Propagate further augment 3D prior bank by sparsifying stored cuboids to mimic far-field proposals;
    OpenBox uses HDBSCAN to perform point cloud cluster and ChatGPT to obtain class-specific cuboid dimensions, and places cuboids on most likely point clusters;
    AnnofreeOD follows Oyster~\citep{zhang2023towards} to run DBSCAN to cluster LiDAR points and L-shape fitting to produce cuboids for the clusters, and applies rigid template matching to decide cuboid orientations.
    Different from these approaches, 
    our auto3D generates instance-specific cuboids for 2D detections, leading to notably better 3D detection performance (\Cref{tab:3D_ablation}).

\end{itemize}

\begin{table}[t]
\centering
\small
\caption{\small
{\bf Analysis of sweep aggregation strategies} on per-class 3D detection performance (mAP$^{3D}$).
We excerpt a few classes but provide all the results in Appx.~\Cref{tab:sweep_aggregation_full_results}.
``$P$+C+$F$'' denotes aggregating the past $P$ sweeps, the current sweep C, and the future $N$ sweeps;
we drop $P$ or $F$ if not aggregating any past or future sweeps.
In each row, we bold the highest number.
Somewhat surprisingly,
aggregation strategies greatly impacts performance on certain classes,
e.g., for {\tt construction}-\texttt{worker}, \texttt{bicycle} and \texttt{traffic}-{\tt cone}, aggregating  future 2 sweeps yields the highest performance gains.
}
\label{tab:sweep_aggregation}
\setlength{\tabcolsep}{2.6mm}
\scalebox{0.86}{
\begin{tabular}{l cccccccccc}
\toprule
 Class & {10+C} & {6+C} & {2+C} & {C} 
 & {C+2} & {C+6} & {C+10} & {1+C+1} 
 & {3+C+3} & {5+C+5} \\
\midrule
{\tt bus}       & 24.3 & 25.8 & 28.1 & \cellcolor{col33}\textbf{30.8} & 28.3 & 27.2 & 26.4 & 29.4 & 26.7 & 25.3 \\
{\tt bicycle}   & 22.5 & 25.1 & 28.6 & 30.1 & \cellcolor{col33}\textbf{32.4} & 29.0 & 26.7 & 30.8 & 29.4 & 28.4 \\
{\tt emergency}-\texttt{vehicle}   & 12.1 &  \cellcolor{col33}\textbf{13.1} & 12.2 & 12.8 &  11.9 &  12.4 &  12.3 & 12.2 &  12.2 & 12.1 \\
{\tt adult} & 34.4 & 43.7 & 56.6 & 59.3 & 60.2 & 46.8 & 36.1 & \cellcolor{col33}\textbf{61.2} & 56.5 & 49.3 \\
{\tt child}   &\ \ 4.4 &\ \  \cellcolor{col33}\textbf{4.9} & \ \ 4.5 & \ \ 3.4 &\ \  2.8 & \ \ 2.6 &\ \  1.9 &\ \  3.5 &\ \  2.9 & \ \ 2.7 \\
{\tt construction}-\texttt{worker}   & 13.6 & 16.3 & 22.9 & 25.6 & \cellcolor{col33}\textbf{28.6} & 24.3 & 20.5 & 27.9 & 25.1 & 22.3 \\
{\tt personal}-\texttt{mobility}   & \ \ 6.6 & \ \ 9.1 & \ \ 8.8 &\ \  8.7 &\ \  9.1 & \ \ 6.9 & \ \ 6.9 & \cellcolor{col33}\textbf{10.4} &\ \  8.6 & \ \ 8.6 \\
{\tt traffic}-\texttt{cone}   & 44.4 & 46.8 & 50.3 & 52.1 & \cellcolor{col33}\textbf{54.1} & 51.4 & 48.6 & 53.3 & 52.1 & 50.2 \\
\bottomrule
\end{tabular}
}
\vspace{-4mm}
\end{table}

{\bf Implementations.}
FMs exploited in this work contain
GroundingDINO as a 2D detector~\citep{liu2023grounding}, 
SAM~\citep{kirillov2023segment} for object segmentation,
and
GPT-4o \citep{achiam2023gpt} for annotation guidelines comprehension
(We also tested Qwen which produced similar results, detailed in Appx.~\S\ref{sec:more-details-2D-appendix}).
We implement all methods with Python and PyTorch on a computing node with four NVIDIA A100 GPUs.
Appx.~\S\ref{sec:open-source-code-experimental-details-appendix}
provides more details.
We release our code in \href{https://github.com/cc50121/AutoExpert-3D-benchmark}{GitHub} and include it in the supplementary material.

\textbf{Benchmarking results.}
As shown in \Cref{tab:3d_results_comparison},
our method auto3D significantly outperforms all the compared methods.
As frustum generation and finetuned GroundingDINO are key components in the pipeline \cref{fig:our_pipeline} and can empower the compared methods,
we apply them to the compared methods (marked by ``w/ frustum'' or ``w/ ft-GD''), leading to significant performance gains.
In particular,
open-vocabulary detectors boosts performance once replacing their off-the-shelf foundational 2D detector with our few-shot finetuned detector (GroundingDINO).
This demonstrates a clear benefit of 2D detector adaptation to expert-crafted guidelines,
corroborating with the previous conclusion \cite{madan2024revisiting} that FMs should be tailored to highly-specialized tasks.

{\bf Analysis on 2D detection.}
\Cref{tab:2D_comparison} compares different prompting and finetuning strategies to improve 2D detection.
Interestingly, changing class names such as \texttt{police}-{\tt officer} to {\tt law} {\tt enforcement} {\tt officer} facilitates FM adaptation (ref. Appx.~\S\ref{sec:more-details-2D-appendix} for all the refined names).
Importantly, finetuning the foundational detector GroundingDINO with refined names performs the best.

\textbf{Analysis of LiDAR sweep aggregation.}
\Cref{tab:sweep_aggregation} reports on some classes detection results by applying different aggregation strategies.
Results show that different aggregations notably improve on certain classes.
Interestingly,
results are ``asymmetric''.
For example, for {\tt traffic-cone}, {\tt construction-worker} and {\tt bicycle}, aggregating the future two sweeps yields significantly better performance than the past two sweeps;
for {\tt child} and {\tt emergency-vehicle}, aggregating the past 6 sweeps is significantly better than others.
We conjecture that this is due to a compound reasons related to rolling shutter and object size and motion pattern of certain classes.

\begin{table}[t]
\centering
\begin{minipage}[c]{0.40\linewidth}
\setlength{\tabcolsep}{1.3mm}
\scalebox{0.85}{
\begin{tabular}{ccccc cc}
\toprule
\multirow{1}{*}{\makecell{\em v-MHT}} & 
\multirow{1}{*}{\makecell{\em SA.}} & 
\multirow{1}{*}{\makecell{$S_{\text{3D}}$}} & 
\multirow{1}{*}{\makecell{\em track}}  & 
 mAP$^{3D}$  & NDS \\
\midrule

\rowcolor{lightgray} & & & & 18.2 & 23.1 \\
\checkmark & & & & 21.9 & 25.2 \\
\checkmark & \checkmark & & & 22.8 & 25.9 \\
\checkmark & \checkmark & \checkmark & & 23.6 & 26.4 \\
\checkmark & \checkmark & \checkmark & \checkmark & {\bf 25.4} & {\bf 27.2} \\
\bottomrule
\end{tabular}
}
\end{minipage}
\hfill
\begin{minipage}[c]{0.58\linewidth}
\vspace{-1.74mm}
\caption{\small
{\bf Ablation study on 3D cuboid generation} (\S\ref{sec:3D-cuboid-search}).
The first row shows results of CM3D~\citep{khurana2024shelf} enhanced by using our finetuned GroundingDINO in 2D detection.
``\emph{SA.}'' uses class-aware sweep aggregation (\Cref{tab:sweep_aggregation});
$S_{\text{3D}}$ incorporates 3D geometric cues to score generated 3D cuboids;
``\emph{track}'' means using 3D tracks to refine scores of generated cuboids.
Clearly, each component contributes with notable performance gains.
}
\label{tab:3D_ablation}
\end{minipage}
\vspace{-3mm}
\end{table}

\textbf{Ablation study in 3D cuboid generation.}
We incrementally include the proposed techniques in our auto3D:
the v-MHT strategy for 3D cuboid generation,
class-aware sweep aggregation, 
geometry-aided scoring, and tracking-based score refinement.
\Cref{tab:3D_ablation} shows that each technique brings 0.8$\sim$3.7 mAP$^{3D}$ gains,
and using them all yields 7.2 mAP$^{3D}$ gains!

\section{Remarks} 
\label{sec:broader-impacts}

{\bf Class imbalance.}
It is well-known that datasets sourced from the real world often exhibit imbalanced data distributions w.r.t. object classes.
Some classes are rare (e.g., {\tt stroller} and {\tt wheelchair}), while some others are quite common (e.g.,  {\tt vehicle} and {\tt pedestrian}).
Consequently, supervised learning methods,
which typically train over massive annotated data,
must handle imbalanced class distributions in the training data~\citep{peri2023towards}.
In contrast, 
AutoExpert itself does not inherently introduce class imbalance,
as expert-crafted annotation guidelines provide similar numbers of visual examples for each defiend object class.
However, as foundation models leveraged to solve AutoExpert are pretrained on massive real-world data which follow imbalanced distributions~\citep{parashar2024neglected}, 
the final methods developed can still inherit data imbalance and exhibit biased predictions.

{\bf Expert-in-the-loop vs. human-in-the-loop.}
AutoExpert does not necessarily mean fully automated annotation without human intervention.
In the contemporary crowd-sourcing annotation paradigm, experts are typically actively involved in annotation procedure as they monitor the quality of human annotations.
Hence, AutoExpert realistically requires ``expert-in-the-loop'', i.e., experts not only design the annotation guidelines but also oversee the progress and quality of annotation.
This is different from human-in-the-loop, which refers to the situation that ordinary human annotators are involved only to label data.

\textbf{Benchmarking foundation models (FMs).}
Although conventional wisdom believes that pretraining on large-scale data will be the key enabler for generalization to  open-world applications, understanding how to appropriately benchmark such methods and pretrained FMs remains challenging. 
FMs have been benchmarked in various ways through general tasks such as reasoning, math, open question answering, and physical rule understanding.
Our AutoExpert benchmarks offer a new venue where various FMs can be assessed w.r.t multiple aspects,
e.g., understanding textual descriptions in annotation guidelines,
summarizing core information from texts and visual examples,
and generalizing to specific object classes for precise detection.
Our benchmarks can facilitate the development of methods for automating data annotation by learning from expert-crafted guidelines.
The developed methods can benefit real-world applications which adopt machine learning solutions,
where data annotation is a prerequisite.
Such applications span industry, health care, and scientific fields.

\begin{figure}[t]
  \centering
  \begin{minipage}[c]{0.60\linewidth}
    \centering
\includegraphics[width=0.95\linewidth, height=1.4in]{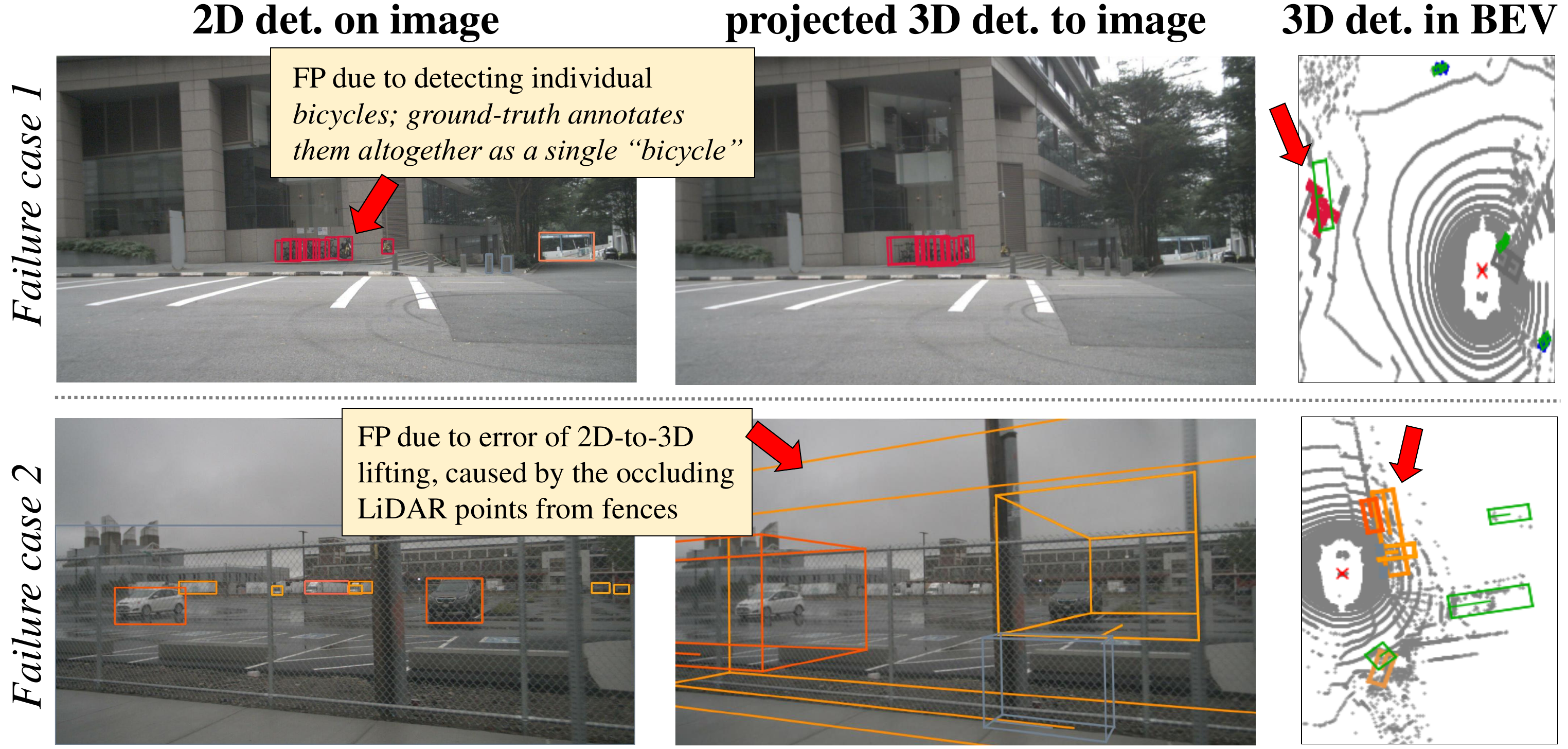}
  \end{minipage}
  \hfill
  \begin{minipage}[c]{0.39\linewidth}
    \small
    \vspace{3mm}
    \caption{\small
    \textbf{Failure cases.}
    We show two failure cases (more in  Appx.~\S\ref{sec:more-visualizations-appendix}).
    (1) The 2D detector produces individual detections in a crowd of bicycles whereas expert annotation requires annotating them altogether. These detections are false positives.
    (2) Although the 2D detector can detect cars and trucks well in the image,
    the occluding fences negatively affects 3D cuboid generation, yielding false positives.
    }
    \label{fig:failure-cases}
  \end{minipage}
  \vspace{-4mm}
\end{figure}

{\bf Broader impact.}
Our work has positive societal impacts.
First,
we observed that annotation guidelines are rarely released with datasets, and Croissant~\citep{NEURIPS2024_9547b09b} does not explicitly mandate their publication. 
Our work advocates updating Croissant to require annotation guidelines, thereby enhancing dataset transparency.
Second, as illustrated by failure cases in \cref{fig:failure-cases},
AutoExpert offers a new venue for evaluating FMs w.r.t their comprehension of annotation guidelines, generalization in detecting nuanced objects in the wild,
and performance on 3D detection tasks in the non-visual LiDAR data modality.
Third, AutoExpert accelerates the methodology development for data auto-annotation,
ultimately contributing to diverse real-world applications in highly specialized domains. 
Nevertheless, 
the techniques presented in this paper could be misused to curate datasets for malicious attacks or introduce demographic biases.
We acknowledge these potential negative societal impacts.

{\bf Limitations and future work.}
First, while we strictly adhere to authentic annotation guidelines that do not provide 3D annotations,
obtaining a few 3D annotations from domain experts is practically feasible.
We investigated this scenario in Appx.~\S\ref{sec:3d-few-shot-appendix}, 
though a comprehensive exploration remains for future work.
Second, while AutoExpert targets unlabeled data which meant to be available during method development, 
our current methods do not leverage such data.
Future work could exploit them via semi-supervised learning to facilitate AutoExpert \cite{liu2026solving}.
Third,
our 3D cuboid generation currently processes object instances in isolation; 
incorporating contextual and proxemic relationships between objects could yield further performance gains.
Fourth,
our work does not attempt to build a LiDAR-based foundation model, a critical yet underexplored direction for future research.

\section{Conclusion}
\label{sec:discussions}

We propose the AutoExpert problem, developing methods for auto-annotation from expert-crafted guidelines.
To enable the exploration of AutoExpert,
we introduce two benchmarks, namely AutoExpert-nuScenes and AutoExpert-PandaSet, by redesigning the evaluation protocol and re-annotating data from the established nuScenes and PandaSet datasets, originally designed for 3D detection research in autonomous driving and having released authentic annotation guidelines.
We address AutoExpert using a conceptually intuitive pipeline,
where we explore a variety of existing methods developed for self-supervised 3D proposal detection and open-vocabulary 3D detection.
To enhance key components in the pipeline, 
we develop several novel techniques, 
including \emph{Multimodal Few-Shot Finetuning} and \emph{VLM-Guided Multi-Hypothesis Testing for 3D cuboid generation}.
Extensive experiments demonstrate that our techniques yield significant performance gains;
yet, AutoExpert remains highly challenging.
We hope this work encourages future research to leverage AutoExpert for evaluating  foundation models and to advance the development of LiDAR-based foundation models.

\section*{Acknowledgements}
This work was supported by 
the National Natural Science Foundation of China (52232015),
the Science and Technology Development Fund of Macau (0058/2025/RIA2, 0067/2024/ITP2), 
the University of Macau (MYRG-CRG2026-00038-FIC, MYRG-GRG2026-00236-FIC, SRG2023-00044-FST),
the Institute of Collaborative Innovation,
and
the CK Foundation, and the Plato Initiative.

{
\small
\bibliographystyle{splncs04}
\bibliography{main}
}

\newpage

\clearpage

{
\begin{center}
\Large
\textbf{AutoExpert: Automating 3D LiDAR Annotation from Expert-Crafted Guidelines}\\(appendix)
\end{center}
}

\renewcommand{\thesection}{\Alph{section}}
\renewcommand{\theHsection}{\Alph{section}}
\setcounter{section}{0}

The appendix supplements our main paper with more details. It is organized as follows: 
\begin{itemize}
\item 
    {\bf Section \ref{sec:more-details-2D-appendix}} provides more implementation details and results of 2D detector finetuning.
\item 
    {\bf Section \ref{sec:lidar-nus-av2-appendix}} studies the performance of transferring a 3D detector supervised trained on a different dataset.
\item 
    {\bf Section \ref{sec:prompt-VL-MHT-appendix}} provides the prompt for virtual expert reasoning for v-MHT.

\item 
    {\bf Section \ref{sec:details-3d-generator-appendix}} provides more details and analyses about v-MHT.

\item 
    {\bf Section \ref{sec:comp_efficiency}} analyzes the computational efficiency of our v-MHT method.
    
\item 
    {\bf Section \ref{sec:supp_occlusion_distance}} analyzes the performance of v-MHT for occluded and far-field objects.

\item 
    {\bf Section \ref{sec:more-details-3d-appendix}} presents detailed results of per-class performance for the compared methods.
    
\item 
    {\bf Section \ref{sec:3d-few-shot-appendix}} studies supervised learning of a 3D cuboid refinement model on some 3D annotations to improve AutoExpert.

\item 
    {\bf Section \ref{sec:sweep_aggregation}} conducts a comparative analysis of various LiDAR sweep aggregation strategies and their impact on detection performance.

\item
    {\bf Section \ref{sec:comparison-3D-2D-tracking-appendix}} shows the superiority of 3D tracking over 2D tracking for score refinement.

\item 
    {\bf Section \ref{sec:pandaset-appendix}} provides details of AutoExpert-PandaSet and benchmarking results of different methods on it.

\item 
    {\bf Section \ref{sec:more-visualizations-appendix}} displays more visual results.
    
\item 
    {\bf Section \ref{sec:image-samples-appendix}} provides image examples available in the expert-crafted annotation guidelines.
    
\item 
    {\bf Section \ref{sec:open-source-code-experimental-details-appendix}} provides open-source code and more experimental details.

\end{itemize}

\section{More Details of  2D Detector Finetuning}
\label{sec:more-details-2D-appendix}

\begin{figure}[t]
    \centering
    \includegraphics[width=0.80\linewidth]{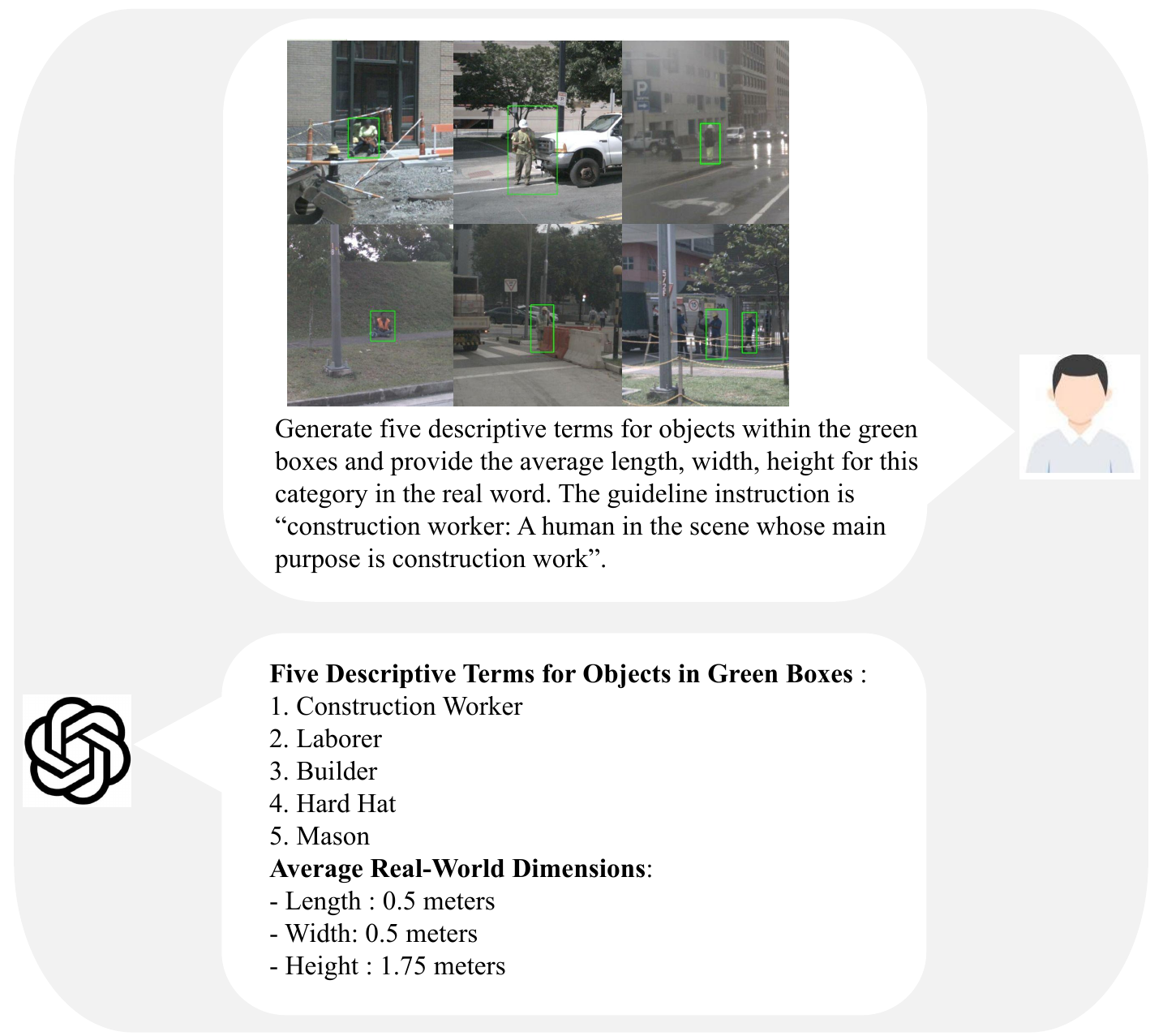}
    \vspace{-1mm}
    \caption{\small
    \textbf{A screenshot of how we search synonyms and object size prior for a given class name.}
    We use both visual examples  and textual description of a specific class available in annotation guidelines.
    Here, we use the {\tt construction-worker} class as an example.
    }
\vspace{-3mm}
\label{fig:ask_gpt4_CW_appendix}
\end{figure}

\textbf{Prompt Refinement to Improve Zero-Shot Detection}.
We refine prompts to improve the zero-shot 2D detection performance with GroundingDINO (\cref{fig:our_pipeline}).
First,
we use GPT-4o to generate five synonyms and object size prior for each object class using the prompt template: 
\textit{``Generate five descriptive terms for objects within the green boxes and provide the average length, width, and height for this category in the real world. The guideline instruction is: [instruction].''}, where \textit{[instruction]} is replaced by actual guideline descriptions.
\cref{fig:ask_gpt4_CW_appendix} displays a screenshot of this step.
Second,
we use each term and their combinations 
to test GroundingDINO's zero-shot 2D detection performance on the validation set.
Third, we select the best term or combination that yields the highest detection precision for each class.
Table \ref{tab:prompt_mapping} summarizes the selected terms for each of the 18 nuScenes classes.
In our work, we also tested using Qwen \citep{qwen} to search for synonyms and object size prior; it produces similar results as GPT-4o, as listed in Table \ref{tab:prompt_mapping} and \Cref{tab:2D_comparison_more}.
This is likely because certain terms are more frequent in the real world for a given class name that different FMs are commonly more familiar with them and prefer to use these frequent terms in a similar way~\citep{parashar2024neglected}.

\begin{table}[t]
\caption{\small
\textbf{Refined class names for the AutoExpert-nuScenes defined classes by GPT-4o \cite{achiam2023gpt} and Qwen \cite{qwen}}.
Notable combinatorial prompts occur in some categories, e.g., the best prompt of {\tt pushable}-\texttt{pullable} is ``pushable pullable garbage container'' and ``hand truck''.
Quantitative comparisons are in \Cref{tab:2D_comparison_more}.
}
\setlength{\tabcolsep}{1.5mm}
\label{tab:prompt_mapping}
\centering
\scalebox{0.75}{
\begin{tabular}{lll}
\toprule
\textbf{original class name} & \textbf{refined class name by GPT-4o} & \textbf{refined class name by Qwen} \\
\midrule
{\tt car} & car & car \\
{\tt truck} & truck & truck \\
{\tt trailer} & trailer, container & trailer, container \\
{\tt bus} & bus & bus \\
{\tt construction}-\texttt{vehicle} & construction-vehicle & construction-vehicle \\
{\tt bicycle} & bicycle bike & bicycle \\
{\tt motorcycle} & narrow motorcycle & motorcycle \\
{\tt emergency}-\texttt{vehicle} & police vehicle, emergency vehicle & police vehicle, ambulance \\
{\tt adult} & adult & adult \\
{\tt child} & single little short youth children & child, kid \\
{\tt police}-\texttt{officer} & law enforcement officer & police-officer, policeman \\
{\tt construction}-\texttt{worker} & construction worker, laborer & construction laborer \\
{\tt stroller} & stroller & stroller \\
{\tt personal}-\texttt{mobility} & personal-mobility, small kick scooter & personal mobility, self-propelled vehicle \\
{\tt pushable}-\texttt{pullable} & pushable pullable garbage container, hand truck & pushable pullable garbage container, hand truck \\
{\tt debris} & debris, full trash bags & debris, full trash bags \\
{\tt traffic}-\texttt{cone} & traffic cone & traffic cone \\
{\tt barrier} & barrier & barrier \\
\bottomrule
\end{tabular}
}
\end{table}


\begin{table}[t]
\centering
\small
\caption{\small
{\bf Comparison of using different refined names for finetuning GroundingDINO on AutoExpert-nuScenes.}
We quantitatively compare performance of finetuned GroundingDINO using the different sets of refined names.
Results are comparable to those in \Cref{tab:2D_comparison}.
showing that both GPT-4o and Qwen  refine class names in the sense of better adapting GroundingDINO to AutoExpert.
}
\vspace{1mm}
\label{tab:2D_comparison_more}
\setlength{\tabcolsep}{5.0mm}
\scalebox{0.85}{
\begin{tabular}{l ccccc}
\toprule
 & 
\makecell{GD\\r-name by GPT-4o} & 
\makecell{GD\\r-name by Qwen} & 
\makecell{ft-GD \\r-name by GPT-4o} &
\makecell{ft-GD \\r-name by Qwen}\\
\midrule
mAP$^{2D}$ & 18.2 & 18.0 & \textbf{20.8} & 20.7 \\
mAP$^{3D}$ & 15.7 & 15.6 & \textbf{18.2} & 18.1 \\
NDS  & 22.1  & 21.9 & \textbf{23.1} & 23.0 \\
\bottomrule
\end{tabular}
}
\vspace{-2mm}
\end{table}

\textbf{Few-Shot Finetuning}.
With the limited amounts of visual examples available in the annotation guidelines, we finetune GroundingDINO.
We test using the original class names and refined names (described in the last paragraph). Refer to the next paragraph for detailed results.
We also adopt data augmentation strategies such as random rotation and cropping. 
Recall that each training image is exclusively annotated with only one  class.
Hence, when finetuning  GroundingDINO,
for each training image, we compute the loss only on the focused class and do not count detections of other classes as false positives.
We use the validation set for model selection and hyperparameter tuning. 
The validation set can be thought of as a simulation of expert intervention in real-world annotation scenarios, where experts are overseeing annotation progress and quality, offering timely intervention when needed.

\textbf{Detailed Results}.
\Cref{tab:2D_comparison} in the main paper summarizes comparisons of using different prompts in the off-the-shelf and finetuned GroundingDINO detectors.
Here, we provide the full results of these methods in Table~\ref{tab:2D_comparison-details-appendix}.
The finetuned GroundingDINO (ft-GD) using refined class names (r-name) achieves significant improvements over the zero-shot baseline ``GD (o-name)'', e.g., on  {\tt child} (from 0.8 to 3.5 in mAP$^{3D}$), {\tt personal}-\texttt{mobility} (from 0.0 to 9.4  in mAP$^{3D}$), {\tt pushable}-\texttt{pullable} (from 1.2 to 4.8 in mAP$^{3D}$), and
{\tt barrier} (from 9.7 to 11.4 in mAP$^{3D}$).

\begin{table}[t]
\centering
\small
\caption{\small
Comparisons of per-category results by  using different finetuning strategies with a  foundational 2D detectors. 
We report the results of zero-shot detector Detic \cite{zhou2022detecting} as a reference, which is used in~\citep{khurana2024shelf}.
}
\label{tab:2D_comparison-details-appendix}
\setlength{\tabcolsep}{1.5mm}
\scalebox{0.83}{
\begin{tabular}{l ccccc ccccc cccc}
\toprule
class
 & 
\multicolumn{2}{c}{Detic (o-name)} & 
\multicolumn{2}{c}{GD (o-name)} & 
\multicolumn{2}{c}{ft-GD (o-name)} & 
\multicolumn{2}{c}{GD (r-name)} & 
\multicolumn{2}{c}{ft-GD (r-name)} \\


\cmidrule(r){2-3}
\cmidrule(r){4-5}
\cmidrule(r){6-7}
\cmidrule(r){8-9}
\cmidrule(r){10-11}

& mAP$^{2D}$ & mAP$^{3D}$
& mAP$^{2D}$ & mAP$^{3D}$
& mAP$^{2D}$ & mAP$^{3D}$
& mAP$^{2D}$ & mAP$^{3D}$
& mAP$^{2D}$ & mAP$^{3D}$\\

\midrule

{\tt car}          & 58.3 & 31.9 & 52.6 & 25.1 &  56.4 & 29.1  & 51.3 & 26.1 & 54.2 & 25.6 \\
{\tt truck}        & 37.2 & 17.6 & 34.3 & 14.2 & 34.4 & 15.0  & 36.3 & 14.1 & 37.5 & 12.6 \\
{\tt trailer}      & 4.2 & 0.8  & 8.5 & 2.0  & 7.0 & 1.3 & 
 8.1 & 1.8   & 8.5 & 1.7   \\
{\tt bus}          & 59.0 & 6.4  & 59.0 & 5.7 & 60.2 & 8.3  & 59.7 & 5.3  & 59.8 & 6.4  \\
{\tt construction}-\texttt{vehicle} & 11.1 & 14.7 & 9.9 & 9.7 & 4.5 & 12.0  & 10.2 & 8.9  & 11.0 & 9.5  \\
{\tt bicycle}      & 28.8 & 28.6 & 22.4 & 22.7 & 28.5 & 27.5 & 22.5 & 19.6 & 24.2 & 29.5 \\
{\tt motorcycle}   & 38.1 & 50.9 & 21.7 & 42.0 & 36.4 & 48.1 & 20.8 & 33.5 & 31.5 & 50.0 \\
{\tt emergency}-\texttt{vehicle}    & 0.5 & 0.2  & 1.6 & 0.9 & 4.3 & 2.4 & 11.9 & 1.5  & 14.2 & 2.6  \\
{\tt adult}        & 10.2 & 5.6  & 23.5 & 54.4 &  36.9 & 53.8 & 23.9 & 53.8 & 31.8 & 58.8 \\
{\tt child}        & 0.0 & 0.00  & 0.9 & 0.8 & 1.1 & 0.7  & 6.6 & 3.2   & 5.2 & 3.5   \\
{\tt police}-\texttt{officer}       & 0.0 & 0.1  & 0.0 & 1.7 &  0.3 & 0.7  & 0.3 & 1.0   & 0.8 & 2.2   \\
{\tt construction}-\texttt{worker} & 0.4 & 2.0  & 5.1 & 27.8 & 9.2 & 28.4  & 4.2 & 28.2  & 5.9 & 25.7  \\
{\tt stroller}     & 1.2 & 7.0  & 13.1 & 27.8 &  13.5 & 20.4 & 14.4 & 29.3       & 15.3 & 21.4       \\
{\tt personal}-\texttt{mobility}  & 0.0 & 0.0  & 0.0 & 0.0 &  0.0 & 0.0 & 0.9 & 0.7       & 10.6 & 9.4       \\
{\tt pushable}-\texttt{pullable}      & 0.0 & 0.0  & 2.6 & 1.2 &  2.9 & 1.1 & 6.1 & 2.8       & 5.8 & 4.8       \\
{\tt debris}     & 0.0 & 0.0  & 0.0 & 0.0 & 0.0 & 0.0 & 0.0 & 0.0       & 0.1 & 0.1       \\
{\tt traffic}-\texttt{cone}    & 51.8 & 50.5  & 46.2 & 44.2 & 52.7 & 40.1  & 46.9 & 43.4       & 52.5 & 52.0       \\
{\tt barrier}     & 0.8 & 0.6  & 3.6 & 9.7 &  2.8 & 8.5 & 2.8 & 8.5       & 5.9 & 11.4      \\
\midrule
mAP$^{3D}$ & 16.5 & 12.1 & 16.9 & 16.1  & 20.0 & 16.6  & 18.2 & 15.7 & \textbf{20.8} & \textbf{18.2} \\

\cmidrule(r){2-3}
\cmidrule(r){4-5}
\cmidrule(r){6-7}
\cmidrule(r){8-9}
\cmidrule(r){10-11}

NDS & \multicolumn{2}{c}{16.6} &  \multicolumn{2}{c}{21.3} & \multicolumn{2}{c}{21.2} & \multicolumn{2}{c}{22.1} & \multicolumn{2}{c}{\textbf{23.1}} \\
\bottomrule
\end{tabular}
}
\end{table}



\section{Cross-Dataset Transfer of a Supervised 3D Detector}
\label{sec:lidar-nus-av2-appendix}

In this work,
we also test a model purposefully trained on another dataset, Argoverse2 (AV2), to investigate whether LiDAR-based pretrained models can generalize to another domain of data captured by a different LiDAR sensor.
Concretely, we train a 3D detector CenterPoint~\citep{yin2021center} on the training set of AV2 in a supervised manner.
Yet, this model exhibits severe performance degradation on the nuScenes benchmark (Table~\ref{tab:AV2_model_transfer}). 
Below, we provide implementation details and analysis on its performance.

\begin{table}[b]
\centering
\small
\caption{\small
\textbf{Performance on AutoExpert-nuScenes  of a model trained on Argoverse2.} A CenterPoint model trained on the AV2 dataset exhibits severe performance degradation when directly evaluated on nuScenes. This demonstrates the challenge for LiDAR-based pretrained models to generalize across domains with different LiDAR sensors.
}
\label{tab:AV2_model_transfer}
\setlength{\tabcolsep}{3.6mm}
\scalebox{0.84}{
\begin{tabular}{llccccccc}
\toprule
Method & pub. & \cellcolor{col11}mAP$^{3D}$ & \cellcolor{col33}NDS & mATE  & mASE & mAOE  & mAVE & mAAE    \\
\midrule

CenterPoint~\citep{yin2021center} & \scriptsize\color{gray}CVPR'21 & \ \  \cellcolor{col11}3.6 & \cellcolor{col33}19.0 & 0.971 & \textbf{0.517} & \textbf{0.794} &  \textbf{0.546} & \textbf{0.447} \\

\textbf{auto3D} & ours  & \cellcolor{col11}\textbf{25.4} & \cellcolor{col33}\textbf{27.2} & \textbf{0.552} &  0.534 & 0.992 & 0.927 & 0.536  \\
\bottomrule
\end{tabular}
}
\vspace{-2.5mm}
\end{table}

{\bf CenterPoint Training on the Argoverse2 Dataset}.
The Argoverse2 (AV2) dataset annotates LiDAR data with 3D cuboids for 30 object classes.
To train a CenterPoint model that can be applied to nuScenes data,
we unify the data format and class vocabulary of  AV2 according to nuScenes.
Then, we supervised-learn the 3D LiDAR detector CenterPoint on the training set of AV2.
After training, we apply it to the AutoExpert test-set. \Cref{tab:AV2_model_transfer} reports its results, showing that CenterPoint significantly underperforms our method.

{\bf Analysis}.
We analyze the LiDAR models of the AV2 and nuScenes,
finding that they have different LiDAR sensor parameters (Table~\ref{tab:lidar_specs}).
As a result, the LiDAR sensors capture data that is different in  (1) measurement range, (2) horizontal/vertical FOV, (3) point cloud density, (4) intensity sensitivity, (5) acquisition frequency.
All these make AV2 LiDAR data different in distribution from the nuScenes LiDAR data,
explaining the poor performance of AV2-trained CenterPoint. 
This further demonstrates the challenge and the need of training LiDAR foundation models.



\begin{table}[t]
\centering
\small
\caption{\small
{\bf Specifics of different LiDAR sensors in nuScenes and Argoverse2 (AV2) for data collection}. 
Clearly, the LiDAR sensors notably differ 
in measurement range, horizontal/vertical FOV, point cloud density, and vertical distribution patterns.
}
\vspace{2mm}
\label{tab:lidar_specs}
\setlength{\tabcolsep}{3.9mm}
\scalebox{0.85}{
\begin{tabular}{lccc}
\toprule
& 
\multicolumn{2}{c}{LiDAR Configuration} \\
\cmidrule(lr){2-3}
parameter/feature & \makecell{nuScenes \citep{caesar2020nuscenes}} & \makecell{AV2 \citep{Argoverse2}} \\ 
\midrule
LiDAR Model & Single HDL-32E & Dual VLP-32C \\
Number of channels & 32 & 32 $\times$ 2 \\
Measurement Range & 
\makecell[c]{Max 100m, Effective 70m} & 
\makecell[c]{Max 200m, Effective 100m} \\
Vertical FOV & \makecell{-30.67$^\circ$ to + 10.67$^\circ$} & \makecell{-25$^\circ$ to + 15$^\circ$} \\
Horizontal FOV & 360$^\circ$ & \makecell{360$^\circ$} \\
Scan Frequency & 20Hz & 10Hz \\
Points per Second & 
\makecell{$\sim$600k pts/s (20Hz $\times$ 30k/frame)} & 
\makecell{$\sim$1.2M pts/s (10Hz $\times$ 120k/frame $\times$ 2)} \\
Vertical Resolution & 
\makecell{0.4$^\circ$ (center), 0.4$^\circ$ to 2.08$^\circ$ (edge)} & 
\makecell{0.33$^\circ$ (center), 1.0$^\circ$ (edge)} \\
Horizontal Resolution & 
0.32$^\circ$ (20Hz) & 
0.096$^\circ$ (10Hz) \\
Intensity Range & \makecell{0-255} & \makecell{0-255} \\
\bottomrule
\end{tabular}
}
\end{table}

\begin{figure}[t]
    \centering
    \includegraphics[width=1.0\linewidth]{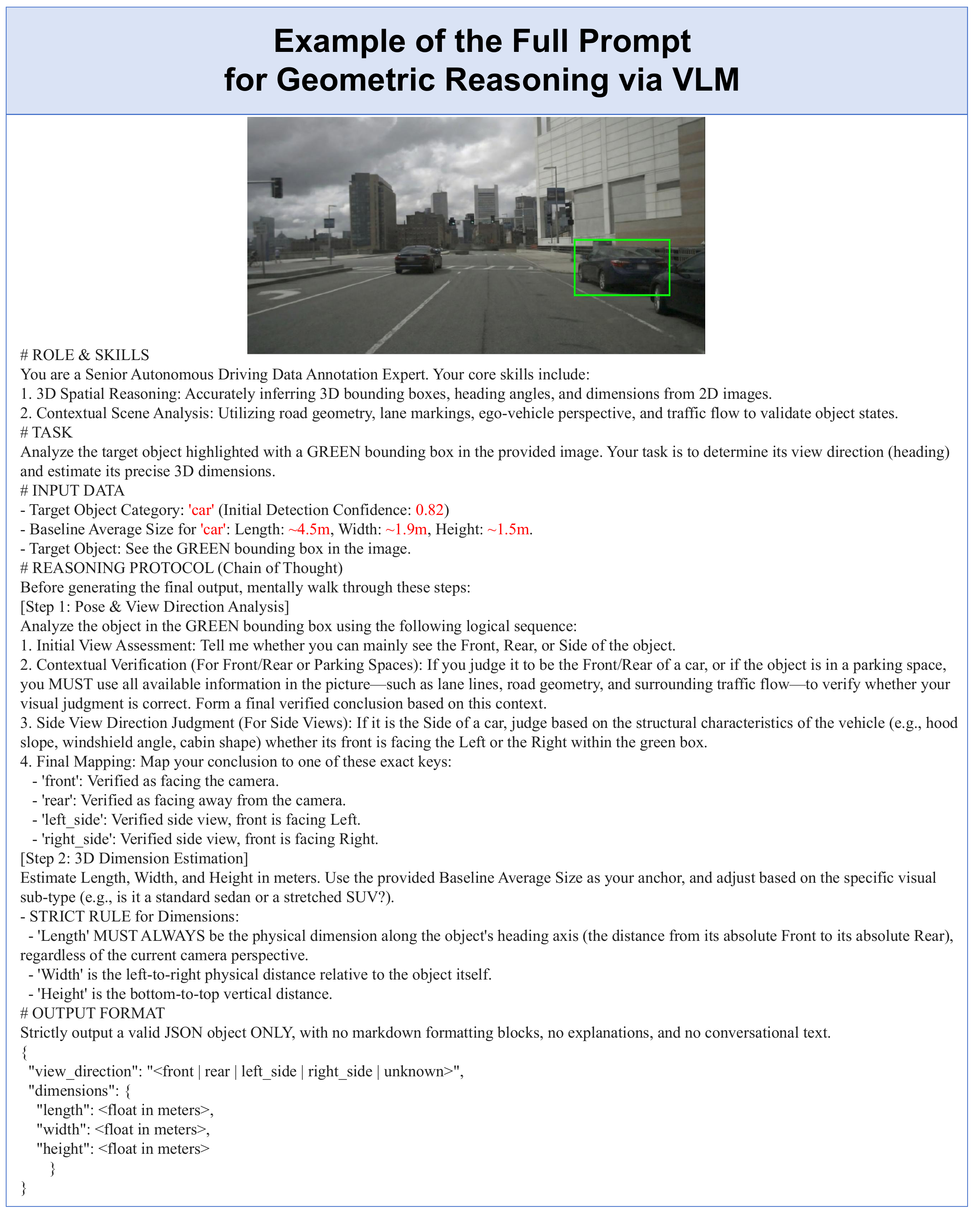}
    \vspace{-2mm}
    \caption{\small
    \textbf{Example of the full prompt used for geometric reasoning via VLM.} The template combines the visual input (image with a green bounding box) with highly structured textual instructions. It guides the Vision Language Model through a Chain-of-Thought reasoning process—leveraging baseline average sizes and contextual cues like road geometry—to output instance-specific 3D dimensions and view directions in a strict JSON format.
    }
\vspace{-3mm}
\label{fig:whole-post-prompt}
\end{figure}

\section{Prompt Details for Virtual Expert Reasoning in v-MHT}
\label{sec:prompt-VL-MHT-appendix}
To fully unlock the reasoning capabilities of Vision Language Model (VLM) for 3D detection, we design a comprehensive and highly structured prompt to instruct VLM to infer physical dimensions and orientations of objects present in images, as illustrated in Figure~\ref{fig:whole-post-prompt}. 
Instead of relying on a simple direct query, our prompt guides the VLM through a Chain-of-Thought (CoT) process to ensure accurate and robust estimation of both the object's orientation and instance-specific dimensions. 
The prompt is structured into four main components:
\begin{itemize}[topsep=1pt]
\item
    \textbf{Role \& Skills:} We initialize the VLM as a ``Senior Autonomous Driving Data Annotation Expert.'' This persona setting encourages the model to leverage its domain-specific knowledge regarding 3D spatial reasoning and contextual scene analysis.
\item    
    \textbf{Input Data:} Alongside the target image where the object is highlighted with a green bounding box, we provide textual priors. Crucially, we supply  the class-specific baseline average size (e.g., length, width, and height for a generic {\tt car}). This acts as a reliable anchor, preventing the VLM from hallucinating unrealistic scales.
\item     
    \textbf{Reasoning Protocol (Chain-of-Thought):} This is the core of our geometric reasoning module. We explicitly instruct the model to perform a step-by-step analysis:
\begin{itemize}
\item  
    Step 1: Pose \& View Direction Analysis. The model first determines the visible faces of the object (Front, Rear, or Side). To mitigate visual ambiguities, we enforce contextual verification, requiring the VLM to analyze lane markings, road geometry, and traffic flow. This step is critical for deriving the estimated orientation $\theta$ and effectively resolving severe orientation ambiguities (e.g., 180-degree heading errors).
    \item  
    Step 2: 3D Dimension Estimation. Instead of uniformly applying the class-average size, the VLM is instructed to adjust the provided baseline dimensions based on the specific visual sub-type (e.g., distinguishing a standard sedan from a stretched SUV). This enables the generation of highly accurate, per-instance dimensions $d$ that better fit the actual object in the image.
\end{itemize}
\item     
    \textbf{Output Format:} Finally, to seamlessly integrate the VLM's output into our automated pipeline, we strictly constrain the response to a valid JSON object, preventing the generation of conversational text and ensuring robust parsing.
\end{itemize}


\begin{table}[t]
    \centering
    \caption{\textbf{Comparison of MHT using per-class cuboid vs. per-instance cuboid.} 
    OpenBox~\citep{lee2025openbox} adopts the former, inquiring GPT to output per-class cuboids prior to place them in 3D towards 3D detection.
    In contrast, our method v-MHT inquires GPT to understand the detected visual object in image and infer the per-instance 3D information.
    This notably helps our method performs better.
    }
    \label{tab:MHT-v-MHT}
    \setlength{\tabcolsep}{4.3mm}
    \scalebox{0.85}{
    \begin{tabular}{l ccccccc}
        \toprule
        & \cellcolor{col11}mAP$^{3D}$  & \cellcolor{col33}NDS & mATE  & mASE  & mAOE  & mAVE & mAAE \\
        \midrule
        MHT  & \cellcolor{col11}19.2 & \cellcolor{col33}23.8 & 0.570 & 0.538 & 1.130 & \textbf{0.920} & 0.555 \\
        v-MHT  & \cellcolor{col11}\textbf{21.9} & \cellcolor{col33}\textbf{25.2} & \textbf{0.560} & \textbf{0.535} & \textbf{0.994} & 0.933 &\textbf{0.549} \\   
        \bottomrule
    \end{tabular}
    }
\end{table}

\section{More Details and Analyses about v-MHT}
\label{sec:details-3d-generator-appendix}

We provide more details of the proposed VLM-Guided Multiple Hypotheses Testing (v-MHT) method for 3D cuboid generation. 
First, for each 2D detection (denoted by $B_{2D}$), instead of relying on generic per-class average dimensions as in standard MHT~\citep{lee2025openbox}, we utilize the VLM to infer the precise \emph{per-instance} 3D cuboid dimensions $d = (l, w, h)$ and an estimated initial orientation $\theta$. We also employ SAM to obtain an accurate instance mask for the object. 
Second, we construct a 3D frustum based on the 2D bounding box and the known camera and LiDAR extrinsic/intrinsic parameters. 
Third, we execute the MHT to refine the cuboid's spatial parameters, denoted as state vector $\mathbf{\Theta} = [x, y, z, \psi]$ (representing the center location and yaw angle). Crucially, instead of blindly searching the full $360^\circ$ rotation space, we leverage the VLM-estimated orientation $\theta$ to establish a reliable initial yaw. This semantic prior effectively resolves the $180^\circ$ orientation ambiguity caused by the geometric symmetry of objects (e.g., confusing the front and rear of a vehicle). Consequently, we constrain the rotation search space to a narrow sector centered around this initial yaw, and define discrete stepsizes for both translation and rotation. 
As quantitatively validated in \Cref{tab:MHT-v-MHT}, replacing the standard MHT's per-class dimensions and unconstrained search with our VLM-guided per-instance priors significantly reduces orientation errors (mAOE) and boosts the overall 3D detection performance.

During the MHT search, for each candidate hypothesis $\mathbf{\Theta}$, we compute the ratio of foreground LiDAR points falling within the generated cuboid $B_{3D}(\mathbf{\Theta})$:
\begin{equation}
\label{eq:coverage}
R(\mathbf{\Theta}) = \frac{1}{|P|}\sum_{p_i \in P} \mathbb{I}(p_i \in B_{3D}(\mathbf{\Theta})),
\end{equation}
where $P$ represents the set of LiDAR points associated with the target object, $p_i$ denotes an individual 3D point within this set, and $\mathbb{I}(\cdot)$ is the indicator function that equals $1$ if the point $p_i$ is located inside the cuboid $B_{3D}(\mathbf{\Theta})$ and $0$ otherwise. Moreover, we calculate the Intersection-over-Union (IoU) between the 2D projection of the 3D cuboid on the image plane, denoted as $\pi(B_{3D}(\mathbf{\Theta}))$, and the original 2D bounding box $B_{2D}$ output by the 2D detector.

Lastly, we select the optimal cuboid parameters $\mathbf{\Theta}^*$ that maximize the joint objective of 3D point coverage and 2D projection alignment:
\begin{equation}
\label{eq:projection}
\mathbf{\Theta}^* = \arg\max_{\mathbf{\Theta}} \Big( R(\mathbf{\Theta}) + \text{IoU}(\pi(B_{3D}(\mathbf{\Theta})), B_{2D}) \Big).
\end{equation}

{\bf Sensitivity analysis.}
We conduct a comprehensive sensitivity analysis of the rotation and translation step size parameters in the proposed MHT-based approach. The analysis aims to determine the optimal parameter values that balance computational efficiency and 3D detection performance.
\Cref{tab:param_sensitivity} demonstrates that our method is robust to a large range of parameter variations.
For rotation step size, values between $\pi/40$ and $\pi/10$ radians yield good and similar performance, with $\pi/10$ radians selected as the default value owing to its favorable balance between computational efficiency and detection accuracy.
Similarly, for translation step size, values between 0.3m and 0.8m maintain good and stable performance, with 0.5m chosen as the default value in our experiments.
Moreover, the sensitivity analysis confirms that coarse step sizes ($\pi/5$ radians for rotation or 1.0m+ for translation) lead to significant performance degradation, while excessively fine step sizes offer diminishing returns and  substantially increased computational costs.

\textbf{Implementation Details for Efficiency.} Generating and evaluating dense hypotheses typically incurs significant computational overhead. To ensure the scalability and efficiency of our v-MHT algorithm, the extensive metric computations—specifically the 3D point-in-box tests and the 2D projection IoU calculations—are heavily optimized. We implement these operations utilizing the Numba compiler~\citep{lam2015numba} combined with GPU parallelization, allowing the search process to evaluate thousands of candidates in a highly parallelized manner with minimal time costs.
\Cref{sec:comp_efficiency} (\Cref{tab:computation_time} therein) details wall-clock time of executing its components.
Please also refer to our released code in \href{https://github.com/cc50121/AutoExpert-3D-benchmark/blob/main/src/masks_to_3D_results.py}{GitHub} or the supplementary material for exact implementations.

\begin{table}[t]
\centering
\caption{\small \textbf{Sensitivity analysis of rotation and translation step size parameters.}
Based on bolded values, we set the corresponding translation and rotation step sizes as default,
which provide good trade-off between computational cost and detection performance.
}
\setlength{\tabcolsep}{6.8mm}
\scalebox{0.85}{
\label{tab:param_sensitivity}
\begin{tabular}{lcccccc}
\toprule
\multicolumn{7}{l}{Rotation Step Size (in radian)} \\
\midrule
Rotation Step & $\pi/40$ & $\pi/30$ & $\pi/20$ & $\pi/10$ & $\pi/5$ \\
mAP$^{3D}$ & 22.0 & 22.0 & 21.9 & \textbf{21.9} & 21.4 \\
NDS & 25.5 & 25.3 & 25.2 & \textbf{25.2} & 24.6 \\
\midrule
\midrule
\multicolumn{7}{l}{Translation Step Size (in meter)} \\
\midrule
Translation Step & 0.3m & 0.5m & 0.8m & 1.0m & 1.5m \\
mAP$^{3D}$  & 22.1 & \textbf{21.9} & 21.3 & 21.0 & 19.1 \\
NDS  & 25.2 & \textbf{25.2} & 24.7 & 24.3 & 22.3 \\
\bottomrule
\end{tabular}
}
\end{table}

\section{Computational Efficiency Analysis}
\label{sec:comp_efficiency}

\begin{table}[b]
\centering
\caption{\small
\textbf{Wall-clock time comparison for processing a single LiDAR sweep.} The evaluation is conducted on a node with four NVIDIA A100 GPUs. By utilizing VLM priors to constrain the geometric search space, the speedup in the MHT phase completely offsets the VLM inference overhead. As a result, our v-MHT achieves the fastest overall processing time.}
\label{tab:computation_time}
\setlength{\tabcolsep}{4.2mm}
\scalebox{0.85}{
\begin{tabular}{l ccccc}
\toprule
Method & \makecell{2D Det. \\ (sec.)} & \makecell{Segmt. \\ (sec.)} & \makecell{VLM Inf. \\ (sec.)} & \makecell{3D Gen. \\ (sec.)} & \makecell{Total \\ Time (sec.)} \\
\midrule
CM3D~\cite{khurana2024shelf}       & 0.08 & 0.01 & N/A  & 0.65 & 0.74 \\
MHT (w/o VLM)          & 0.08 & 0.01 & N/A  & 0.86 & 0.95 \\
\textbf{v-MHT (Ours)}              & 0.08 & 0.01 & 0.40 & 0.15 & \textbf{0.64} \\
\bottomrule
\end{tabular}
}
\end{table}

One may intuitively think that our v-MHT approach is computationally expensive due to the introduction of Large Vision Language Models and the iterative multiple hypotheses testing. However, our designed hybrid search strategy and hardware-optimized implementation ensure that our pipeline is not only highly scalable but also more efficient than existing baselines.
To optimize the computational overhead, we introduce a confidence-aware routing mechanism based on the 2D detection scores from GroundingDINO. Specifically, for detected objects with a confidence score above $0.3$, we deploy the VLM to infer the instance-specific dimensions and initial orientation. This strong semantic prior allows the subsequent MHT to perform a highly constrained search within a remarkably narrow sector. Conversely, for objects with a confidence score below the $0.3$ threshold, we bypass the VLM and fall back to the traditional MHT, which utilizes class-average dimensions and performs a global $360^\circ$ search to recover potentially difficult objects.

We conduct the runtime analysis on a compute node equipped with four NVIDIA A100 GPUs and compare it against the recent work CM3D~\cite{khurana2024shelf}. The averaged wall-clock time per LiDAR sweep on the AutoExpert test set is reported in \Cref{tab:computation_time}. To ensure a strictly fair comparison, both methods leverage the upgraded hardware for the shared foundational modules. As a result, the execution times for GroundingDINO (2D detection) and SAM (foreground segmentation) are significantly accelerated to 0.08 and 0.01 seconds, respectively.
Crucially, \Cref{tab:computation_time} demonstrates the efficiency advantage of our v-MHT. For a comprehensive analysis, we also evaluate a variant of our method using only the traditional MHT (without VLM priors), which relies on an exhaustive $360^\circ$ geometric search. While the batched VLM inference across the four A100 GPUs adds an overhead of approximately 0.40 seconds, this is more than offset by the massive reduction in the MHT search space. Guided by the strong semantic priors from the VLM and further accelerated by Numba-compiled GPU parallelization, our 3D generation phase takes merely 0.15 seconds. 
In contrast, CM3D requires 0.65 seconds for its 3D generation, and our traditional MHT variant requires a heavy 0.86 seconds for global search. Consequently, our v-MHT achieves a total processing time of just 0.64 seconds per sweep, substantially outperforming both the traditional MHT variant (0.95 seconds) and the CM3D baseline (0.74 seconds). This highlights a compelling trade-off: injecting high-level semantic reasoning via VLMs effectively accelerates the downstream geometric optimization, achieving both superior accuracy and competitive efficiency for automated data annotation.


\begin{table}[t]
\centering
\caption{\small \textbf{Analysis of occlusion robustness (mAR$^{3D}$) and distance performance (mAP$^{3D}$).} Our v-MHT method consistently outperforms CM3D, with particularly significant gains for heavily occluded objects (0-20\% visibility) and distant targets (20-30m). The final method incorporating LiDAR aggregation, 3D cuboid scoring with geometric cues and tracking-based refinement demonstrates substantial improvements, especially for far-field and occluded objects.}
\setlength{\tabcolsep}{3.3mm}
\scalebox{0.80}{
\begin{tabular}{lcccccccc}
\toprule
 \multirow{2}{*}{Method} & \multicolumn{4}{c}{\textbf{mAR$^{3D}$/Occlusion Level}} & \multicolumn{4}{c}{\textbf{mAP$^{3D}$/Distance (m)}} \\
\cmidrule(r){2-5} \cmidrule(r){6-9}
 & 60-100\% & 40-60\% & 20-40\% & 0-20\% & 0-10 & 10-20 & 20-30 & 0-50 \\
\midrule
CM3D~\citep{khurana2024shelf} & 33.5\% & 49.4\% & 58.1\% & 59.9\% & 26.2 & 23.9 & 13.9 & 18.2 \\
\midrule
\textbf{v-MHT (ours)} & 34.1\% & 51.5\% & 59.6\% & 61.9\% & 30.2 & 26.2 & 15.6 & 21.9 \\
 & (+0.6\%) & (+2.1\%) & (+1.5\%) & (+2.0\%) & (+4.0) & (+2.3) & (+1.7) & (+3.7) \\
\midrule
\textbf{auto3D} & \textbf{36.5\%} & \textbf{54.0\%} & \textbf{60.8\%} & \textbf{63.1\%} & \textbf{30.4} & \textbf{29.8} & \textbf{19.8} & \textbf{25.4} \\
 & (+3.0\%) & (+4.6\%) & (+2.7\%) & (+3.2\%) & (+4.2) & (+5.9) & (+5.9) & (+7.2) \\
\bottomrule
\end{tabular}
}
\label{tab:occlusion_distance_analysis}
\end{table}

\section{3D Cuboid Generation for Occluded and Far-Field Objects by v-MHT}
\label{sec:supp_occlusion_distance}

We analyze the performance of our v-MHT 3D cuboid generation method for occluded and far-field objects.
We provide quantitative evaluations using nuScenes' visibility tags and distance stratification on the AutoExpert test set. 
As our method and previous approaches such as CM3D do not predict occlusion levels, computing Average Precision (AP) for occlusion analysis is not appropriate. 
Instead, we report mean Average Recall (mAR$^{3D}$), following the nuScenes protocol to average over distance thresholds \{0.5, 1.0, 2.0, 4.0\} meters across all 18 classes. 
For distance analysis, we report mAP$^{3D}$ averaged over these thresholds. 

Table~\ref{tab:occlusion_distance_analysis} provides breakdown results of our method auto3D and the compared CM3D.
Our method consistently outperforms CM3D across all occlusion levels and distance ranges.
Importantly, our final method, which incorporates LiDAR aggregation, 3D cuboid scoring with geometric cues and tracking-based refinement, yields particularly large performance gains for heavily occluded objects (0-20\% visibility) and distant targets (20-30m). These improvements can be attributed to the densification of LiDAR points through aggregation and refinement techniques, which significantly aid in detecting far-field small objects and occluded targets. This analysis validates the robustness of our approach in challenging scenarios involving occlusion and long distance.

\begin{table}[t]
\centering
\small
\caption{\small \textbf{Per-class 3D object detection performance (Part 1 of 2).} Comparison of our proposed \textbf{auto3D} against baseline methods on standard vehicle and adult pedestrian categories. Best results are highlighted in \textbf{bold}. Here, {\tt CV} and {\tt EV} denote the \texttt{construction\_vehicle} and \texttt{emergency\_vehicle} classes, respectively.}
\label{tab:3d_results_comparison_part1}
\setlength{\tabcolsep}{3.5mm}
\scalebox{0.85}{
\begin{tabular}{l ccc ccc cc}
\toprule
Method  & {\tt car} & {\tt truck} & {\tt trailer}  & {\tt bus} & {\tt CV}  & {\tt bicycle} & {\tt motorcycle}  & {\tt EV}\\
\midrule
SAM3D~\citep{zhang2023sam3d} & 6.2 & 5.2 & 0.2 & 2.1 & 0.5 & 3.3 & 4.1 & 0.1\\
Oyster~\citep{zhang2023towards} & 13.1 & 4.1 & 0.4 & 2.7 & 3.1 & 10.0 & 17.1 & 0.8\\
\quad w/ frustum  & 20.1 & 9.0 & 1.2 & 4.6 & 6.8 & 21.1 & 35.7 & 1.9\\

LISO~\citep{baur2024liso} & 19.0 & 7.0 & 0.8 & 6.4 & 4.3 & 13.7 & 23.1 & 2.5\\
\quad w/ frustum  & 25.0 & 12.8 & 1.4 & 11.7 & 7.9 & 25.1 & 42.3 & 4.5\\

UNION~\citep{lentsch2024union} & 18.8 & 7.2 & 0.6 & 6.6 & 4.8 & 14.2 & 22.8 & 2.4\\
\quad w/ frustum  & 29.2 & 13.1 & 1.6 & 13.5 & 8.0 & 26.1 & 43.4 & 4.8\\

CPD~\citep{wu2024commonsense} & 20.7 & 7.7 & 0.9 & 7.2 & 4.6 & 14.4 & 22.6 & 2.5\\
\quad w/ frustum  & 26.5 & 13.9 & 1.6 & 12.9 & 8.3 & 27.8 & 42.7 & 4.5\\

\midrule

Find\&Prop~\citep{etchegaray2024find} & 24.3 & 8.6 & 0.2 & 11.1 & 4.1 & 45.1 & 35.8 & 2.1\\
\quad  w/ ft-GD & 36.5 & 12.9 & 0.3 & 16.6 & 6.2 & \textbf{57.7} & \textbf{53.7} & 3.2\\

OpenSight~\citep{zhang2024opensight} & 24.1 & 8.3 & 0.8 & 5.1 & 6.1 & 24.1 & 28.2 & 1.6\\
\quad  w/ ft-GD & 35.9 & 12.4 & 1.2 & 7.6 & 9.1 & 35.9 & 42.0 & 2.4\\

Openbox~\citep{lee2025openbox} & 29.9 & 9.0 & 0.8 & 6.2 & 5.7 & 14.0 & 21.5 & 5.5\\
\quad  w/ ft-GD & 29.4 & 17.1 & 1.2 & 12.9 & 9.2 & 20.4 & 30.7 & 7.3\\

AnnofreeOD~\citep{sun2025annofreeod} & 32.2 & 8.8 & 0.9 & 14.5 & 7.9 & 14.9 & 23.5 & 6.0\\
\quad  w/ ft-GD & 32.7 & 16.2 & 1.2 & 21.6 & 10.6 & 23.0 & 34.5 & 9.2\\

CM3D~\citep{khurana2024shelf} & 31.9 & 17.6 & 0.8 & 6.4 & 14.7 & 28.6 & 50.9 & 0.2\\
\quad  w/ ft-GD & 31.0 & 12.6 & 1.7 & 6.4 & 9.5 & 29.5 & 50.0 & 2.6\\

\midrule
\textbf{auto3D} & \textbf{43.9} & \textbf{23.6} & \textbf{2.0} & \textbf{31.4} & \textbf{16.0} & 33.4 & 51.2 & \textbf{13.1} \\
\bottomrule
\end{tabular}
}
\vspace{-1mm}
\end{table}

\begin{table}[t]
\centering
\small
\caption{\small \textbf{Per-class 3D object detection performance (Part 2 of 2).} Continuation of Table~\ref{tab:3d_results_comparison_part1}, presenting the evaluation on specialized pedestrian types, personal mobility devices, and static road elements. Best results are highlighted in \textbf{bold}. {\tt PO}, {\tt PM}, {\tt PP}, and {\tt TC} denote \texttt{police\_officer}, \texttt{personal\_mobility}, \texttt{pushable\_pullable}, and \texttt{traffic\_cone}, respectively.}
\label{tab:3d_results_comparison_part2}
\setlength{\tabcolsep}{2.9mm}
\scalebox{0.85}{
\begin{tabular}{l ccc ccc ccc c}
\toprule
Method  &  adult & {\tt child} & {\tt PO} & {\tt CW}  & {\tt stroller} & {\tt PM}  & {\tt PP} & {\tt debris}  & {\tt TC} & {\tt barrier}\\
\midrule
SAM3D~\citep{zhang2023sam3d} & 0.9 & 0.1 & 0.2 & 0.2 & 1.5 & 0.9 & 0.1 & 0.1 & 1.6 & 1.5\\
Oyster~\citep{zhang2023towards} & 19.8 & 1.2 & 0.9 & 8.6 & 6.9 & 3.7 & 0.1 & 0.1 & 16.9 & 3.9\\
\quad w/ frustum  & 42.0 & 2.5 & 1.6 & 18.4 & 15.3 & 6.7 & 0.1 & 0.1 & 37.1 & 8.1\\

LISO~\citep{baur2024liso} & 26.7 & 1.6 & 1.0 & 11.7 & 9.8 & 4.0 & 0.1 & 0.1 & 23.6 & 5.2\\
\quad w/ frustum  & 47.2 & 2.9 & 1.8 & 21.4 & 17.9 & 7.3 & 0.1 & 0.1 & 43.3 & 9.6\\

UNION~\citep{lentsch2024union} & 25.9 & 1.7 & 1.0 & 12.0 & 9.6 & 3.8 & 0.1 & 0.1 & 38.1 & 4.9\\
\quad w/ frustum  & 50.6 & 3.5 & 2.6 & 25.2 & 16.9 & 7.0 & 0.1 & 0.1 & 46.7 & 10.0\\

CPD~\citep{wu2024commonsense} & 27.9 & 2.5 & 1.6 & 13.3 & 13.1 & 5.9 & 0.1 & 0.1 & 27.6 & 9.0\\
\quad w/ frustum  & 52.3 & 4.5 & 3.0 & 25.9 & 27.3 & 10.7 & 4.3 & 0.1 & 41.8 & 14.1\\

\midrule

Find\&Prop~\citep{etchegaray2024find} & 19.2 & 0.5 & 0.6 & 6.9 & 12.3 & 2.9 & 3.9 & 0.1 & 21.3 & 4.4\\
\quad  w/ ft-GD & 23.8 & 0.8 & 0.9 & 10.4 & 13.5 & 4.4 & \textbf{5.9} & 0.2 & 52.4 & 6.6\\

OpenSight~\citep{zhang2024opensight} & 32.6 & 0.9 & 0.7 & 7.8 & 18.1 & 4.2 & 1.6 & 0.1 & 25.6 & 22.5\\
\quad  w/ ft-GD & 48.6 & 1.3 & 1.0 & 11.6 & 27.0 & 6.3 & 2.4 & \textbf{0.2} & 38.1 & \textbf{33.5}\\

Openbox~\citep{lee2025openbox} & 26.8 &  2.2 & 1.3 & 13.0 & 14.7 & 5.4 & 2.2 & 0.1 & 26.1 & 7.0\\
\quad  w/ ft-GD & 38.0 & 3.7 & 2.2 & 18.1 & 28.1 & 7.7 & 3.2 & 0.1 & 34.9 & 9.9\\

AnnofreeOD~\citep{sun2025annofreeod} & 19.6 & 2.0 & 1.7 & 14.3 & 21.6 & 5.4 & 2.9 & 0.1 & 26.0 & 8.0\\
\quad  w/ ft-GD & 40.0 & 3.5 & 2.3 & 21.3 & 31.4 & 9.2 & 3.6 & 0.1 & 39.2 & 11.0\\

CM3D~\citep{khurana2024shelf} & 5.6 &  0.0 & 0.1 & 2.0 & 7.0 & 0.0 & 0.0 & 0.0 & 50.5 & 0.6\\
\quad  w/ ft-GD & 58.8 & 3.5 & 2.2 & 25.7 & 21.4 & 9.4 & 0.1 & 0.1 & 52.0 & 11.4\\

\midrule
\textbf{auto3D} & \textbf{63.3} &\textbf{5.4} & \textbf{3.6} & \textbf{31.1} & \textbf{47.0} & \textbf{12.8} & 5.3 & 0.1 & \textbf{57.4} & 16.6  \\
\bottomrule
\end{tabular}
}
\vspace{-1mm}
\end{table}

\section{Detailed Per-Class Evaluation Results}
\label{sec:more-details-3d-appendix}

Due to space limitations in the main paper, we provide the comprehensive per-class evaluation results of our proposed auto3D framework and the baseline methods in this supplementary document. The detailed performance breakdown across all evaluated categories is split into two parts: Table~\ref{tab:3d_results_comparison_part1}\&\ref{tab:3d_results_comparison_part2}.
Table~\ref{tab:3d_results_comparison_part1} reports the detection performance on standard vehicle categories and common active road users. Our auto3D demonstrates substantial improvements over prior state-of-the-art methods across the majority of these dominant classes. Notably, it achieves significant performance gains on the {\tt car} (43.9), {\tt truck} (23.6). This underscores the effectiveness of our VLM-guided geometric reasoning in accurately estimating dimensions and resolving orientation ambiguities for common, structurally diverse objects.
Table~\ref{tab:3d_results_comparison_part2} extends the evaluation to specialized pedestrian types, personal mobility devices, and static road elements. These categories often pose extreme challenges for 3D detection due to their long-tail distributions, small scales, or irregular geometries. Even in these highly challenging scenarios, auto3D maintains a dominant performance advantage. For instance, it establishes new state-of-the-art results on the {\tt stroller} (47.0), \texttt{traffic\_cone} ({\tt TC}, 57.4), and \texttt{construction\_worker} ({\tt CW}, 31.1) categories. 
Overall, the consistent superiority of auto3D across both frequently occurring vehicles and rare, long-tail categories highlights its robust generalization capabilities. It further validates that combining foundational 2D perception with VLM-guided 3D hypotheses testing is a highly effective paradigm for open-vocabulary 3D scene understanding.

 \begin{figure}[t]
    \centering
    \includegraphics[trim=0cm 0 0cm 0cm, clip, width=0.7\linewidth]{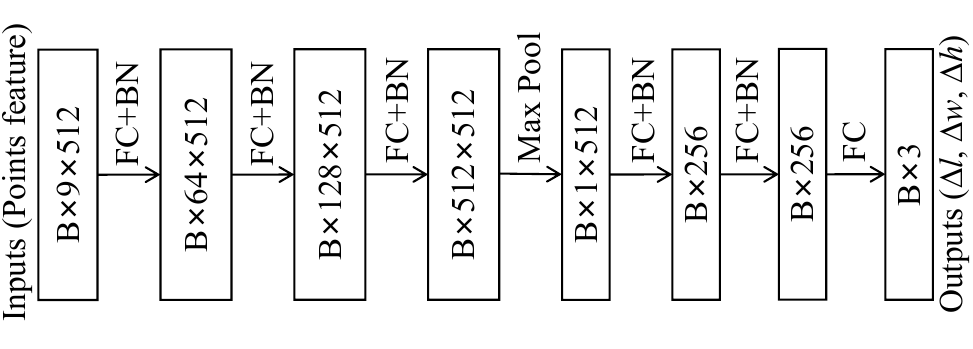}
    \vspace{-4mm}
    \caption{\small
\textbf{Architecture of the proposed PointNet~\citep{qi2017pointnet} model $P_\phi$ for 3D cuboid refinement}. 
Here, $\mathrm{B}$ denotes the batch size, $\mathrm{FC}$ corresponds to a fully connected layer, and $\mathrm{BN}$ represents a batch normalization layer. 
The input to the model consists of 512 LiDAR points, each represented by a 9-dimensional feature vector (as defined in Equation~\ref{eq:point_input_feature}). 
If the number of points in the frustum point cloud is fewer than 512, random oversampling is applied to reach the target count of 512 points. 
Conversely, if the point cloud contains more than 512 points, random downsampling is performed to reduce the number to 512. 
The model outputs dimensional offsets used to refine the size produced by our v-MHT 3D cuboid generation method.
}
\vspace{-1mm}
\label{fig:3d-few-shot-newtwork-appendix}
\end{figure}

\section{Few-Shot Supervised Learning for 3D Refinement }
\label{sec:3d-few-shot-appendix}

In the supplement,
we explore whether training on a small amount of annotated LiDAR data helps 3D detection atop our method auto3D.
Specifically, we assume that the LiDAR data corresponding to the few-shot visual images available in annotation guidelines are annotated with 3D cuboids.
Then, we study whether training a lightweight model on such annotated LiDAR data can refine 3D detections.
We design a PointNet network $P_\phi$~\citep{qi2017pointnet}, as shown in \cref{fig:3d-few-shot-newtwork-appendix},
which takes as input the 3D locations of LiDAR points and outputs 3D cuboid dimension's offset based on the 3D detection by auto3D.
We optionally train the network to output orientation offset, confidence score, and cuboid center offset.

To prepare such training data, we apply our finetuned GroundingDINO, SAM, and the v-MHT 3D cuboid generation method to the limited amount of training images.
Hence, for each 2D detection, we obtain its corresponding object mask, frustum, and 3D cuboid dimension $\mathbf{d}_0=[l_0; w_0; h_0]$, its center location $[x_0; y_0; z_0]$ and orientation $\theta_0$. 

For this 2D detection, we have the input data to the PointNet as the set of 3D locations of LiDAR points from the corresponding frustum.
Importantly, we transform these 3D locations (say $\mathbf{p}_{\text{trans}} = [x; y; z]$) depending on the 3D cuboid center location and orientation:
\begin{equation}
\mathbf{p}_{\text{trans}} = \begin{bmatrix}
\cos\theta_0 & \sin\theta_0 & 0 \\
-\sin\theta_0 & \cos\theta_0 & 0 \\
0 & 0 & 1
\end{bmatrix} \left( 
\begin{bmatrix}x\\y\\z\end{bmatrix}
 - \begin{bmatrix}x_0\\y_0\\z_0\end{bmatrix} \right).
\end{equation}

With the transformed coordinates of LiDAR points, for each point, 
we construct a 9-dim feature $\mathbf{f}$ using the following equation:
\begin{equation}
\mathbf{f} = \big[ 
\mathbf{p}_{\text{trans}};
\mathbf{d}_0 - \mathbf{p}_{\text{trans}}; 
\mathbf{d}_0 + \mathbf{p}_{\text{trans}} \big]
\label{eq:point_input_feature}
\end{equation}



The network $P_\phi$ outputs an offset $\Delta d$ between the ground-truth dimension $l_{\text{gt}},w_{\text{gt}},h_{\text{gt}}$ and the dimension of the initial 3D cuboid in log scale:
\begin{equation}
\small
\label{eq:offset}
\Delta l = \log\left(\frac{l_{\text{gt}}}{l_0}\right), \ \
\Delta w = \log\left(\frac{w_{\text{gt}}}{w_0}\right), \ \ 
\Delta h = \log\left(\frac{h_{\text{gt}}}{h_0}\right)
\end{equation}

When training the network, we adopt a smooth L1 loss with the default $\beta$ hyperparameter 1.0.

\Cref{tab:3d-few-shot} lists the results by training the model to output different offsets.
Despite careful tuning of the network architecture and hyperparameters, 
the model yields only a small improvement of 1.5 NDS by being trained to output size offset.
We believe that the scarcity of 3D labeled data poses significant challenges for training a generalizable 3D perception model.
The results, together with \Cref{tab:3d_results_comparison}, 
suggest a need for developing LiDAR-based foundation models.

\begin{table}[t]
\centering
\caption{\small
{\bf Analysis of learning to refine generated 3D cuboids.}
Suppose we manually prepare 3D cuboids on the LiDAR point clouds for the visual examples provided in the annotation guidelines.
We use them to learn a model that takes as input a generated 3D cuboid and outputs refined cuboid. 
The refinement can translate the cuboid through re-\emph{centering}, adjust cuboid \emph{size}, tune the \emph{orientation}, and re-\emph{score} the cuboid.
We tune the model over the validation set.
Results show that 
learning such a refinement network on few-shot annotated LiDAR data is challenging,
although 
learning to refine the size of generated cuboids on limited examples is beneficial.
}
\label{tab:3d-few-shot}
\setlength{\tabcolsep}{3.1mm}
\scalebox{0.81}{
\begin{tabular}{ccccccccccc}
\toprule
{\em center} & \emph{size} & \emph{orientation} & \emph{score} & \cellcolor{col11}mAP$^{3D}$  & \cellcolor{col33}NDS & mATE  & mASE  & mAOE  & mAVE & mAAE    \\
\midrule
& & & & \cellcolor{col11}25.4 & \cellcolor{col33}27.2 & 0.552 & 0.534 & 1.133 & 0.927 & 0.536  \\
$\checkmark$ & & & &  \cellcolor{col11}24.5 & \cellcolor{col33}24.7 & 0.601 & 0.589 & 1.142 & 0.976 & 0.585  \\
 & $\checkmark$ & &  & \cellcolor{col11}25.4 & \cellcolor{col33}{\bf 28.7} & 0.552 & 0.386 & 1.113 & 0.927 & 0.536 \\
 & & $\checkmark$ & & \cellcolor{col11}25.4 & \cellcolor{col33}27.7 & 0.552 & 0.492 & 0.990 & 0.927 & 0.536  \\
 & $\checkmark$ & $\checkmark$ & & \cellcolor{col11}25.4 & \cellcolor{col33}28.0 & 0.552 & 0.460 & 1.111 & 0.927 & 0.536  \\
  & & & $\checkmark$ & \cellcolor{col11}23.8 & \cellcolor{col33}23.8 & 0.613 & 0.600 & 1.241 & 0.982 & 0.614  \\
\bottomrule
\end{tabular}
}
\vspace{-3mm}
\end{table}

\begin{table}[t]
\centering
\small
\caption{\small
{\bf Analysis of sweep aggregation strategies on per-class 3D detection performance} (mAP$^{3D}$).
``$P$+C+$F$'' denotes aggregating the past $P$ sweeps, the current sweep $C$, and the future $N$ sweeps;
we drop $P$ or $F$ if not aggregating any past or future sweeps.
In each row, we bold the highest number and highlight it if exceeding other numbers by 0.5 points.
Somewhat surprisingly,
aggregation strategies greatly impacts performance on certain classes,
e.g., {\tt construction}-\texttt{worker}, \texttt{bicycle} and \texttt{traffic}-{\tt cone}, aggregating the past 2 sweeps yields remarkably better performance than other strategies.
}
\vspace{0mm}
\label{tab:sweep_aggregation_full_results}
\setlength{\tabcolsep}{3.2mm}
\scalebox{0.80}{
\begin{tabular}{l cccccccccc}
\toprule
 Class & {10+C} & {6+C} & {2+C} & {C} 
 & {C+2} & {C+6} & {C+10} & {1+C+1} 
 & {3+C+3} & {5+C+5} \\
\midrule
{\tt car} & 33.6 & 36.4 & 39.9 & \cellcolor{col33}\textbf{41.6} & 40.8 & 38.2 & 35.5 & 40.7 & 38.6 & 37.4 \\
{\tt truck} & 20.7 & 21.6 & 22.7 & 22.7 & 22.8 & 22.3 & 21.4 & 22.9 & 22.7 & 22.2 \\
{\tt trailer} & 1.7 & 1.7 & 1.8 & 1.7 & 1.7 & 1.7 & 1.7 & 1.8 & 1.7 & 1.7 \\
{\tt bus} & 24.3 & 25.8 & 28.1 & \cellcolor{col33}\textbf{30.8} & 28.3 & 27.2 & 26.4 & 29.4 & 26.7 & 25.3 \\
{\tt construction}-\texttt{vehicle} & 13.5 & 14.3 & 14.6 & 14.9 & 14.9 & 14.4 & 14.3 & 14.8 & 14.8 & 14.6 \\
{\tt bicycle} & 22.5 & 25.1 & 28.6 & 30.1 & \cellcolor{col33}\textbf{32.4} & 29.0 & 26.7 & 30.8 & 29.4 & 28.4 \\
{\tt motorcycle} & 37.6 & 42.2 & 48.5 & 50.7 & 49.6 & 43.6 & 38.0 & \cellcolor{col33}\textbf{51.2} & 48.7 & 45.5 \\
{\tt emergency}-\texttt{vehicle} & 12.1 & 13.1 & 12.2 & 12.8 & 11.9 & 12.4 & 12.2 & 13.1 & 12.2 & 12.1 \\
{\tt adult} & 34.4 & 43.7 & 56.6 & 59.3 & 60.2 & 46.8 & 36.1 & \cellcolor{col33}\textbf{61.2} & 56.5 & 49.3 \\
{\tt child} & 4.4 & 4.9 & 4.5 & 3.4 & 2.8 & 2.6 & 1.9 & 3.5 & 2.9 & 2.7 \\
{\tt police}-\texttt{officer}
& 1.2 & 1.5 & 2.2 & 2.2 & 2.2 & 2.0 & 1.8 & 2.3 & 2.0 & 1.9 \\
{\tt construction}-\texttt{worker} & 13.6 & 16.3 & 22.9 & 25.7 & \cellcolor{col33}\textbf{28.6} & 24.3 & 20.5 & 27.9 & 25.1 & 22.3 \\
{\tt stroller} & 19.2 & 20.2 & 21.5 & 21.5 & 23.4 & 24.1 & 24.2 & 21.7 & 21.5 & 23.2 \\
{\tt personal}-\texttt{mobility} & 6.6 & 9.1 & 8.8 & 8.7 & 9.1 & 7.0 & 6.9 & \cellcolor{col33}\textbf{10.4} & 8.6 & 8.6 \\
{\tt pushable}-\texttt{pullable} & 4.6 & 4.7 & 4.6 & 4.8 & 5.0 & 4.8 & 4.8 & 4.8 & 4.8 & 4.8 \\
{\tt debris} & 0.1 & 0.1 & 0.1 & 0.1 & 0.1 & 0.1 & 0.1 & 0.1 & 0.1 & 0.1 \\
{\tt traffic}-\texttt{cone} & 44.4 & 46.8 & 50.3 & 52.1 & \cellcolor{col33}\textbf{54.1} & 51.4 & 48.6 & 53.3 & 52.1 & 50.2 \\
{\tt barrier} & 11.2 & 11.2 & 11.3 & 11.4 & 11.4 & 11.4 & 11.4 & 11.4 & 11.4 & 11.4 \\
\midrule
\textbf{mAP} & 17.0 & 18.8 & 21.1 & 21.9 & 22.2 & 20.2 & 18.5 & \cellcolor{col33}\textbf{22.2} & 21.1 & 20.1 \\
\textbf{NDS} & 24.5 & 24.9 & 24.9 & 25.2 & 25.4 & 25.1 & 24.5 & \cellcolor{col33}\textbf{25.4} & 25.1 & 25.0 \\
\bottomrule
\end{tabular}
}
\vspace{-2mm}
\end{table}

\section{LiDAR Sweep Aggregation Strategies}
\label{sec:sweep_aggregation}

\Cref{tab:sweep_aggregation_full_results} presents per-class $mAP^{3D}$ performance by using different sweep aggregation strategies.
The systematical evaluation of these aggregation strategies reveals optimal aggregation configurations tailored to specific object classes.

Furthermore, 
we also tested per-instance LiDAR sweep aggregation (per-instance SA) \cite{baur2024liso, wu2024commonsense, zhang2023towards}, 
which aggregates multiple sweeps if the detected object is static determined by its 3D tracking trajectory.
\Cref{tab:sweep_aggregation_diff_methods} compares its performance against the baseline that does not adopt sweep aggregation and our per-class sweep aggregation strategy (per-class SA).
The per-instance SA improves over the baseline
but underperforms our per-class SA.
We conjecture the reasons are three-fold.
First, estimating static-vs-moving by per-instance SA is erroneous that has limited densification of LiDAR points for significantly improvement.
Second, dynamic objects still have sparse points in per-instance SA that make them difficult to detect.
Third,
the follow-up step by v-MHT for 3D cuboid generation offers robustness to smears of per-class SA, 
as it uses fixed instance-specific cuboids (to search for a 3D placement as the generated 3D cuboid) that effectively mitigate the effect of smears.

\begin{table}[t]
    \centering
    \caption{\small
    \textbf{Results of LiDAR sweep aggregation methods}.
    The baseline does not adopt LiDAR sweep aggregation (SA).
    We compare the method that aggregates LiDAR sweeps for static objects (named as per-instance SA) and our per-class SA.
    The per-instance SA improves over the baseline
but underperforms our per-class SA.
We conjecture the reasons are two-fold:
(1) estimating static-vs-moving by per-instance SA is erroneous that has limited densification of LiDAR points for significantly improvement,
(2) dynamic objects still have sparse points that make them difficult to detect,
and (3) the follow-up step by v-MHT for 3D cuboid generation offers robustness, as it uses fixed instance-specific cuboids (to search for a 3D placement as the generated 3D cuboid) that effectively mitigate the effect caused by smears in aggregated sweeps.
    }
    \label{tab:sweep_aggregation_diff_methods}
    \setlength{\tabcolsep}{4.3mm}
    \scalebox{0.85}{
    \begin{tabular}{l ccccccc}
        \toprule
        & \cellcolor{col11}mAP$^{3D}$  & \cellcolor{col33}NDS & mATE  & mASE  & mAOE  & mAVE & mAAE \\
        \midrule
        baseline w/o SA  & \cellcolor{col11}21.9 & \cellcolor{col33}25.2 & 0.560 & 0.535 & 0.994 & 0.933 & 0.549 \\
        per-instance SA  & \cellcolor{col11}22.1 & \cellcolor{col33}25.4 & 0.559 & 0.537 & 0.994 & 0.928 & 0.544 \\
        per-class SA & \cellcolor{col11}\textbf{22.8} & \cellcolor{col33}\textbf{25.9} & \textbf{0.556} & \textbf{0.536} & \textbf{0.993} & \textbf{0.919} & \textbf{0.538} \\        
        \bottomrule
    \end{tabular}
    }
\end{table}

\begin{figure}[t]
    \centering
    \includegraphics[width=1.0\linewidth]{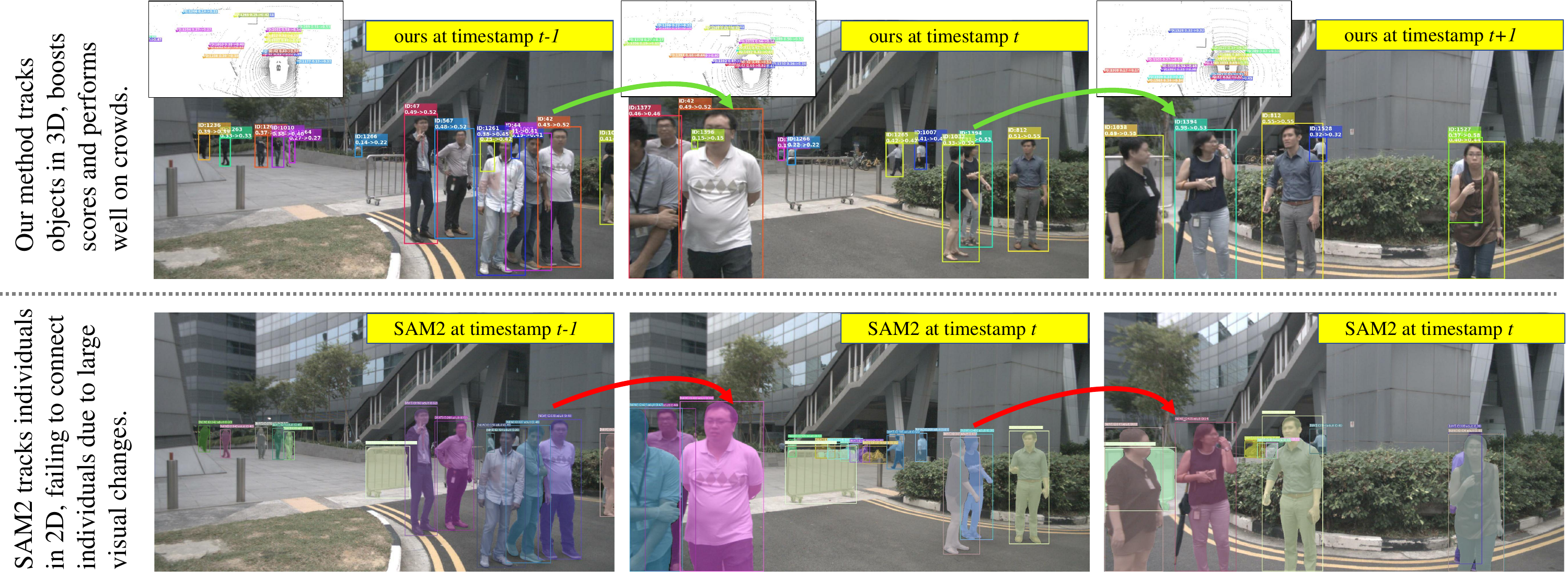}
    \vspace{-5mm}
    \caption{\small
    \textbf{Qualitative comparison of tracking-based score refinement strategies.} \textbf{Top:} Our proposed method performs tracking in the 3D space (highlighted by the point cloud insets). It robustly maintains object identities (green arrows) across consecutive timestamps ($t-1$, $t$, $t+1$) despite significant scale changes in crowded scenes, enabling effective score boosting. \textbf{Bottom:} Relying solely on 2D visual tracking via SAM2 fails to connect the same individuals (red arrows). The severe 2D visual variations and occlusions lead to fragmented tracks, rendering temporal score refinement ineffective.
    }
\vspace{-0mm}
\label{fig:tracking-2D-3D}
\end{figure}

\section{3D Tracking vs. 2D Tracking for Score Refinement}
\label{sec:comparison-3D-2D-tracking-appendix}

As discussed in Section \ref{ssec:tech-endeavor} of the main paper, we introduce a tracking-based score refinement module to enhance the temporal consistency and confidence of our detections. An intuitive alternative to our 3D tracking approach would be to utilize advanced 2D video segmentation and tracking foundational models, such as SAM2~\cite{ravi2024sam}, to track instances directly in the 2D image plane prior to score refinement. In this section, we provide a qualitative visual comparison to demonstrate why our 3D-centric tracking approach is strictly necessary and superior for this task.

Figure~\ref{fig:tracking-2D-3D} illustrates a challenging crowded pedestrian scenario across consecutive timestamps ($t-1$, $t$, and $t+1$). When relying solely on SAM2 for 2D tracking (shown in the bottom row), the model struggles to maintain consistent instance identities temporally. As pedestrians move relative to the ego-vehicle, they undergo drastic 2D visual changes in scale, appearance, and perspective. Furthermore, crowded scenes introduce severe mutual occlusions in the camera view. These significant visual variations cause the 2D tracker to frequently lose target association (indicated by the red arrows), treating an existing person as a newly appeared object. Consequently, the temporal chain is broken, making it impossible to effectively aggregate confidence scores across frames.
Beyond temporal inconsistencies, 2D tracking introduces severe spatial limitations in modern autonomous driving setups. A standard LiDAR frame is typically accompanied by a surround-view system consisting of multiple cameras (e.g., 6 cameras in nuScenes). Foundational trackers like SAM2 inherently operate within a single camera view. To achieve holistic scene tracking across the entire ego-vehicle surroundings, relying on 2D tracking would mandate an additional, complex cross-camera Re-Identification (Re-ID) module to associate the same instance across overlapping camera fields of view at the exact same timestamp. This mandatory cross-camera matching inevitably incurs further accuracy degradation (due to severe viewpoint and illumination changes across cameras) and significant computational overhead.

In contrast, our proposed method (shown in the top row) tracks objects directly in the unified 3D space. By operating in the 3D world coordinate system, our approach naturally sidesteps the cumbersome cross-camera Re-ID problem entirely, as the multi-view visual inputs are already grounded into a single geometric space. Moreover, the physical locations, velocities, and dimensions of objects in 3D change continuously and predictably—free from 2D perspective distortions and severe scale variations. As highlighted by the 3D point cloud insets, this spatial continuity enables our system to robustly connect the same individuals (indicated by the green arrows) even in highly dense crowds. This robust spatial-temporal association serves as the fundamental cornerstone that allows our method to successfully boost the detection scores and suppress false negatives efficiently.

\section{Benchmark Results on AutoExpert-PandaSet}
\label{sec:pandaset-appendix}

\begin{figure}[t]
    \centering
    \includegraphics[width=1.0\linewidth]{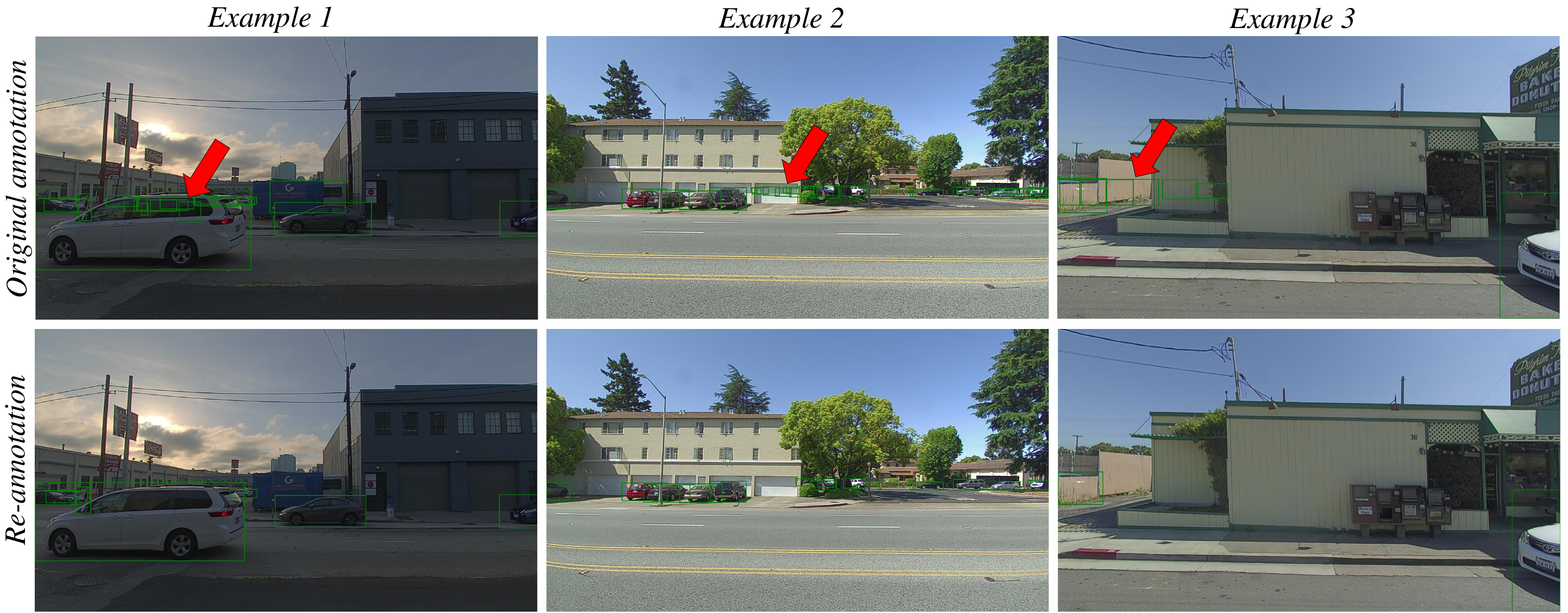}
    \vspace{-5mm}
    \caption{\small
    \textbf{Flawed annotations in the original PandaSet and our re-annotations for the AutoExpert-PandaSet benchmark.} 
    Clearly, the original PandaSet annotations (ref. the top row) correspond to ``ghost'' objects that are entirely invisible.
    Our re-annotations address this issue (ref. the bottom row).}
    \label{fig:reannotation_comparison}
\end{figure}


\begin{table*}[t]
\centering
\caption{\small \textbf{Distribution of our re-annotation per category on AutoExpert-PandaSet.} 
We sample 200 non-continuous frames and meticulously re-annotate a total of 4,695 instances across 25 diverse categories that are visible in the 2D images.}
\label{tab:pandaset_classes}
\setlength{\tabcolsep}{15.8mm}
\scalebox{0.88}{
\begin{tabular}{lr}
\toprule
\textbf{Class Name} & \textbf{GT Count} \\
\midrule
\texttt{car} & 2,464 \\
\texttt{pedestrian} & 890 \\
\texttt{pylons} & 263 \\
\texttt{temporary\_construction\_barriers} & 179 \\
\texttt{pedestrian\_with\_object} & 122 \\
\texttt{cones} & 111 \\
\texttt{pickup\_truck} & 106 \\
\texttt{bicycle} & 101 \\
\texttt{medium-sized\_truck} & 96 \\
\texttt{road\_barriers} & 62 \\
\texttt{signs} & 56 \\
\texttt{bus} & 49 \\
\texttt{motorcycle} & 47 \\
\texttt{other\_vehicle-uncommon} & 28 \\
\texttt{rolling\_containers} & 26 \\
\texttt{construction\_signs} & 18 \\
\texttt{animals-other} & 17 \\
\texttt{other\_vehicle-pedicab} & 16 \\
\texttt{other\_vehicle-construction\_vehicle} & 15 \\
\texttt{tram\_or\_subway} & 9 \\
\texttt{emergency\_vehicle} & 5 \\
\texttt{motorized\_scooter} & 4 \\
\texttt{personal\_mobility\_device} & 4 \\
\texttt{towed\_object} & 4 \\
\texttt{semi-truck} & 3 \\
\midrule
\textbf{Overall} & \textbf{4,695} \\
\bottomrule
\end{tabular}
}
\end{table*}

\begin{table}[t]
\caption{\small
\textbf{Refined class names for the AutoExpert-Pandaset defined classes by GPT-4o \cite{achiam2023gpt}}.
Quantitative comparisons are in \Cref{tab:2D_comparison_more_pandaset}.
}
\setlength{\tabcolsep}{8.8mm}
\label{tab:prompt_mapping_PandaSet}
\centering
\scalebox{0.85}{
\begin{tabular}{ll}
\toprule
\textbf{original class name} & \textbf{refined class name by GPT-4o} \\
\midrule
\texttt{car} & \texttt{car} \\
\texttt{pedestrian} & \texttt{pedestrian} \\
\texttt{pylons} & \texttt{pylons} \\
\texttt{temporary\_construction\_barriers} & \texttt{plastic\_pedestrian\_barricade} \\
\texttt{pedestrian\_with\_object} & \texttt{pedestrian\_with\_open\_umbrella} \\
\texttt{cones} & \texttt{tall\_safety\_cone} \\
\texttt{pickup\_truck} & \texttt{heavy-duty\_pickup\_truck} \\
\texttt{bicycle} & \texttt{bicycle} \\
\texttt{medium-sized\_truck} & \texttt{box\_truck} \\
\texttt{road\_barriers} & \texttt{barrier} \\
\texttt{signs} & \texttt{stop\_sign\_assembly} \\
\texttt{bus} & \texttt{rigid\_three-axle\_bus} \\
\texttt{motorcycle} & \texttt{motorcycle} \\
\texttt{other\_vehicle-uncommon} & \texttt{rickshaw\_van} \\
\texttt{rolling\_containers} & \texttt{wheeled\_garbage\_toter} \\
\texttt{construction\_signs} & \texttt{construction\_signs} \\
\texttt{animals-other} & \texttt{animals} \\
\texttt{other\_vehicle-pedicab} & \texttt{auto\_rickshaw} \\
\texttt{other\_vehicle-construction\_vehicle} & \texttt{other\_vehicle-construction\_vehicle} \\
\texttt{tram\_or\_subway} & \texttt{tram} \\
\texttt{emergency\_vehicle} & \texttt{fire\_engine} \\
\texttt{motorized\_scooter} & \texttt{stand-up\_electric\_scooter} \\
\texttt{personal\_mobility\_device} & \texttt{kick\_scooter} \\
\texttt{towed\_object} & \texttt{utility\_trailer} \\
\texttt{semi-truck} & \texttt{semi-truck\_tractor} \\
\bottomrule
\end{tabular}
}
\end{table}

\begin{table}[t]
\centering
\small
\caption{\small
{\bf Comparison of 2D detectors w.r.t 2D and 3D metrics on AutoExpert-PandaSet.}
We test the performance of prompting or finetuning GroundingDINO (GD) with original class names vs. refined names (r-name) given by the GPT-4o model (\Cref{tab:prompt_mapping_PandaSet}). Results show that finetuning GD (ft-GD) using r-name performs the best.
}
\vspace{1mm}
\label{tab:2D_comparison_more_pandaset}
\setlength{\tabcolsep}{12.0mm}
\scalebox{0.85}{
\begin{tabular}{l cccc}
\toprule
 & mAP$^{2D}$ & mAP$^{3D}$ & NDS \\
 \midrule
GD o-name & 17.3 & 10.1 & 17.7 \\
ft-GD o-name & 25.9 & 17.9 & 26.1 \\
GD r-name & 20.5  & 13.6 & 20.9 \\
ft-GD r-name & \textbf{27.8}  & \textbf{18.4} & \textbf{27.6} \\
\bottomrule
\end{tabular}
}
\vspace{-2mm}
\end{table}

We now provide more details about the AutoExpert-PandaSet benchmark,
which builds upon the PandaSet dataset~\cite{xiao2021pandaset}.
PandaSet provides authentic expert-crafted guidelines and defines 25 object categories.

{\bf Data re-annotation.}
We observe significant misalignment and flaws in the original 3D annotations of PandaSet:
many 3D annotations correspond to invisible objects, which are completely invisible in the RGB images, as illustrated in \cref{fig:reannotation_comparison}.
Therefore, we re-annotate a subset of PandaSet data with high-quality well-aligned 3D boxes.
Our re-annotated data select 200 discrete (non-continuous) keyframes, where contain 4,695 cuboids in total.
Table~\ref{tab:pandaset_classes} lists the number of annotations for each of the 25 classes.

\begin{table}[t]
\centering
\caption{\small \textbf{Benchmarking results on the 
AutoExpert-PandaSet test-set.}
Evaluated across 25 diverse categories on 200 frames. Due to the lack of temporal and attribute data, the NDS metric is adapted. Our auto3D significantly outperforms OpenBox, AnnofreeOD and CM3D.}
\label{tab:pandaset-results}
\setlength{\tabcolsep}{6.0mm}
\scalebox{0.83}{
\begin{tabular}{lcccccc}
\toprule

Method & pub. & \cellcolor{col11}mAP$^{3D}$ & \cellcolor{col33}NDS & mATE  & mASE & mAOE   \\

\midrule

OpenBox \citep{lee2025openbox} & \scriptsize\color{gray}NeurIPS'25 & \cellcolor{col11}10.9 & \cellcolor{col33}15.0 & 0.856 & 0.491 & 1.185  \\
\quad  w/ ft-GD   & ours & \cellcolor{col11}15.9 & \cellcolor{col33}18.9 & 0.820 & 0.461 & 1.076  \\

AnnofreeOD \citep{sun2025annofreeod} & \scriptsize\color{gray}ICCV'25 & \cellcolor{col11}11.5 & \cellcolor{col33}15.4 & 0.855 & 0.486 & 1.190  \\
\quad  w/ ft-GD   & ours & \cellcolor{col11}16.1 & \cellcolor{col33}19.3 & 0.811 & 0.452 & 1.015  \\

CM3D \citep{khurana2024shelf} & \scriptsize\color{gray}CoRL'24 & \cellcolor{col11}12.3 & \cellcolor{col33}16.0 & 0.831 & 0.522 & 1.182  \\
\quad  w/ ft-GD   & ours & \cellcolor{col11}16.6 & \cellcolor{col33}19.7 & 0.803 & 0.455 & 1.104  \\

\midrule

\textbf{auto3D} & ours & \cellcolor{col11}\textbf{18.4} & \cellcolor{col33}\textbf{27.6} & \textbf{0.775} & \textbf{0.376} & \textbf{0.802}\\
\bottomrule
\end{tabular}
}
\vspace{-3mm}
\end{table}

\begin{table}[t]
\centering
\caption{\small \textbf{Per-Class results by our auto3D on the AutoExpert-PandaSet  test-set.}
The abbreviations used in the table are \texttt{Const\_Signs} for Construction Signs; \texttt{Emerg\_Vehicle} for Emergency Vehicle; \texttt{Med\_Truck} for Medium-sized Truck; \texttt{Mot\_Scooter} for Motorized Scooter; \texttt{Oth\_Veh} for Other Vehicle; \texttt{Ped\_w/\_Object} for Pedestrian with Object; \texttt{Pers\_Mob\_Device} for Personal Mobility Device; \texttt{Roll\_Containers} for Rolling Containers; \texttt{Temp\_Const\_Barr} for Temporary Construction Barriers; \texttt{Tram/Subway} for Tram or Subway.}
\label{tab:per-class-results}
\setlength{\tabcolsep}{8.0mm}
\scalebox{0.83}{
\begin{tabular}{lccccc}
\toprule
Class Name &\cellcolor{col11}mAP$^{3D}$ & \cellcolor{col33}NDS & ATE  & ASE & AOE \\
\midrule
\texttt{Animals-Other}                &\cellcolor{col11}36.4 &  \cellcolor{col33}48.5 & 0.513 & 1.084 & 1.358 \\
\texttt{Bicycle}                      &\cellcolor{col11}37.4 &  \cellcolor{col33}31.3 & 0.539 & 0.296 & 0.743 \\
\texttt{Bus}                          &\cellcolor{col11}17.1 &  \cellcolor{col33}3.6 & 2.011 & 0.257 & 0.552 \\
\texttt{Car}                          &\cellcolor{col11}40.8 &  \cellcolor{col33}35.6 & 0.846 & 0.058 & 0.610 \\
\texttt{Cones}                        &\cellcolor{col11}41.4 &  \cellcolor{col33}34.3 & 0.345 & 0.163 & 0.900 \\
\texttt{Const\_Signs}                 &\cellcolor{col11}16.6 &  \cellcolor{col33}7.4 & 0.583 & 0.661 & 0.803 \\
\texttt{Emerg\_Vehicle}               &\cellcolor{col11}52.7 &  \cellcolor{col33}37.5 & 0.441 & 0.134 & 0.088 \\
\texttt{Med\_Truck}                   &\cellcolor{col11}14.2 &  \cellcolor{col33}2.9 & 1.295 & 0.251 & 0.757 \\
\texttt{Motorcycle}                   &\cellcolor{col11}33.2 &  \cellcolor{col33}20.2 & 0.419 & 0.142 & 0.799 \\
\texttt{Mot\_Scooter}                 &\cellcolor{col11}15.5 &  \cellcolor{col33}0.7 & 0.288 & 0.507 & 1.787 \\
\texttt{Oth\_Veh-Const}               &\cellcolor{col11}21.7 &  \cellcolor{col33}12.9 & 0.909 & 0.324 & 0.679 \\
\texttt{Oth\_Veh-Pedicab}             &\cellcolor{col11}41.4 &  \cellcolor{col33}32.8 & 0.542 & 0.124 & 0.659 \\
\texttt{Oth\_Veh-Uncommon}            &\cellcolor{col11}21.8 &  \cellcolor{col33}5.5 & 0.507 & 0.371 & 0.652 \\
\texttt{Pedestrian}                   &\cellcolor{col11}37.8 &  \cellcolor{col33}30.9 & 0.357 & 0.160 & 1.065 \\
\texttt{Ped\_w/\_Object}              &\cellcolor{col11}30.9 &  \cellcolor{col33}22.7 & 0.452 & 0.209 & 1.139 \\
\texttt{Pers\_Mob\_Device}            &\cellcolor{col11}35.9 &  \cellcolor{col33}18.2 & 0.228 & 0.215 & 0.598 \\
\texttt{Pickup\_Truck}                &\cellcolor{col11}33.3 &  \cellcolor{col33}22.0 & 0.650 & 0.088 & 0.697 \\
\texttt{Pylons}                       &\cellcolor{col11}27.9 &  \cellcolor{col33}23.6 & 0.384 & 0.627 & 0.940 \\
\texttt{Road\_Barriers}               &\cellcolor{col11}26.0 &  \cellcolor{col33}8.7 & 0.630 & 0.243 & 0.483 \\
\texttt{Roll\_Containers}             &\cellcolor{col11}26.6 &  \cellcolor{col33}14.8 & 0.396 & 0.288 & 0.924 \\
\texttt{Semi-truck}                   &\cellcolor{col11}0.0 &  \cellcolor{col33}0.0 & 1.000 & 1.000 & 1.000 \\
\texttt{Signs}                        &\cellcolor{col11}20.4 &  \cellcolor{col33}5.0 & 0.373 & 0.584 & 0.663 \\
\texttt{Temp\_Const\_Barr}            &\cellcolor{col11}35.0 &  \cellcolor{col33}38.5 & 0.462 & 0.665 & 1.023 \\
\texttt{Towed\_Object}                &\cellcolor{col11}6.1 &  \cellcolor{col33}0.6 & 2.025 & 0.649 & 0.894 \\
\texttt{Tram/Subway}                  &\cellcolor{col11}18.8 &  \cellcolor{col33}0.9 & 3.188 & 0.302 & 0.240 \\
\bottomrule
\end{tabular}%
}
\vspace{-3mm}
\end{table}

\textbf{Dataset availability and evaluation server.} 
We have not released our re-annotated PandaSet data but built \href{https://huggingface.co/spaces/autoexpert-cvpr2026-workshop/auto3D}{an evaluation serve on HuggingFace} for a hosted challenge competition.
The dataset of our AutoExpert-PandaSet is available at the \href{https://huggingface.co/datasets/autoexpert-cvpr2026-workshop/dataset-public}{HuggingFace repository}.
After the challenge, we will release our annotations on test-set to foster research on AutoExpert.

\textbf{Evaluation metrics.} 
We adopt mAP$^{3D}$ and NDS metrics as used in the AutoExpert-nuScenes benchmark.
Notably, as our AutoExpert problem focuses on 3D detection,
we do not incorporate the averaged Attribute Error (AAE) and mean Velocity Error (AVE) in NDS.
Therefore, we re-define the NDS metric as below:
\begin{equation}\small
\text{NDS} = \frac{1}{8} \Big[ 5 \times \text{mAP} + (1 - \min(1, \text{ATE}))  + (1 - \min(1, \text{ASE})) + (1 - \min(1, \text{AOE})) \Big]
\end{equation}

\textbf{Results.} 
\Cref{tab:prompt_mapping_PandaSet} lists the refined names by GPT-4o on the AutoExpert-PandaSet benchmark.
\Cref{tab:2D_comparison_more_pandaset} confirms that finetuning GroundingDINO on refined names (found by GPT-4o) performs better than other choices, corrobrating the conclusion on the AutoExpert-nuScenes benchmark (Tables~\ref{tab:2D_comparison} and \ref{tab:2D_comparison_more}).
Table~\ref{tab:pandaset-results} compares our auto3D method and recent methods including OpenBox, AnnoFreeOD, and CM3D.
Our auto3D significantly outperforms the compared methods:
it achieves an $\text{mAP}^{3D}$ of 18.4 and an NDS of 27.6, surpassing CM3D by $+6.1$ and $+11.6$, respectively.
The results re-confirm the robustness and generalizability of our method.
Furthermore, the per-class evaluation detailed in Table~\ref{tab:per-class-results} demonstrates auto3D's strong capability across diverse categories, achieving notably high NDS scores on critical classes such as {\tt Emergency}-\texttt{Vehicles} (52.7), {\tt Cones} (41.4), and \texttt{Cars} (40.8).

\section{More Visualizations}
\label{sec:more-visualizations-appendix}

\begin{figure*}[t]
\centering
\includegraphics[width=0.999\linewidth]{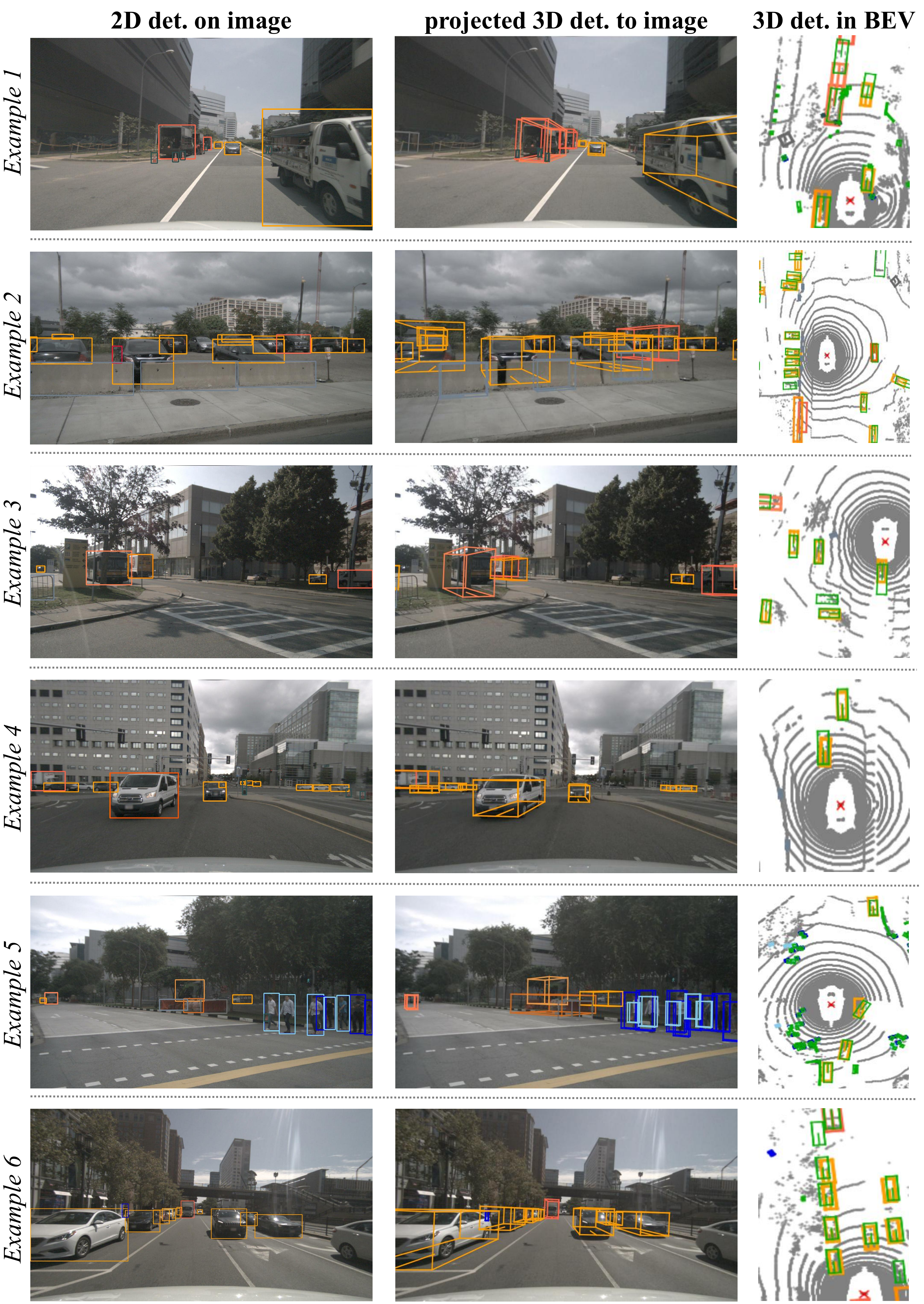}
\vspace{-1mm}
\caption{
More visual results of 2D detection and generated 3D cuboids (i.e., 3D detection)  of auto3D on AutoExpert-nuScenes.
}
\label{fig:results_visual_more}
\end{figure*}

\begin{figure*}[t]
\centering
\includegraphics[width=0.9998\linewidth]{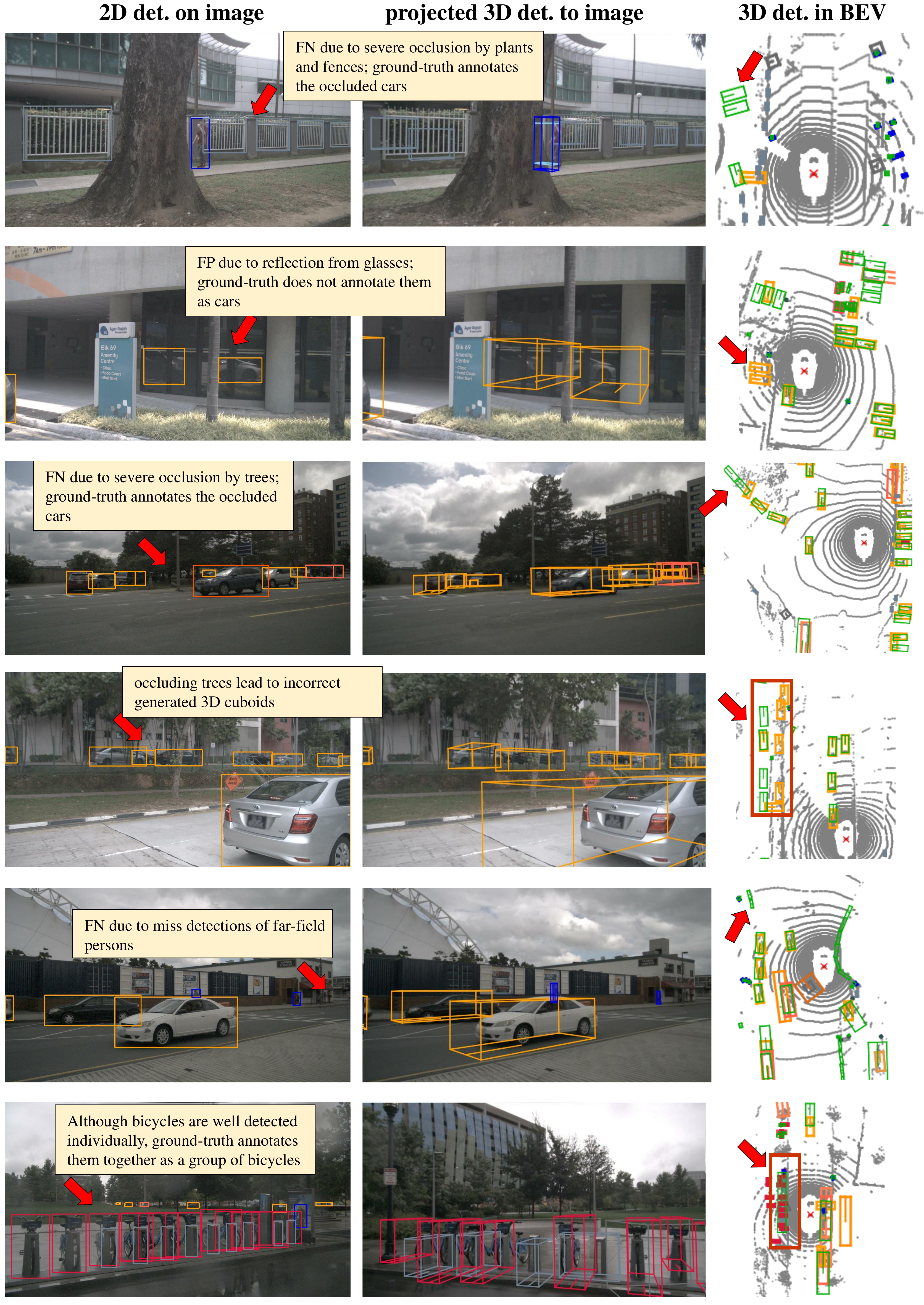}
\vspace{-1mm}
\caption{
Failure cases of auto3D on AutoExpert-nuScenes.
}
\label{fig:results_visual_more_failure_cases}
\end{figure*}

Fig.~\ref{fig:results_visual_more} presents more visual results on the AUtoExpert-nuScenes benchmark.
\cref{fig:results_visual_more_failure_cases} presents more failure cases.

\section{Visual Demonstrations in Annotation Guidelines}
\label{sec:image-samples-appendix}

\begin{figure*}[t]
\centering
\includegraphics[width=1.0\linewidth]{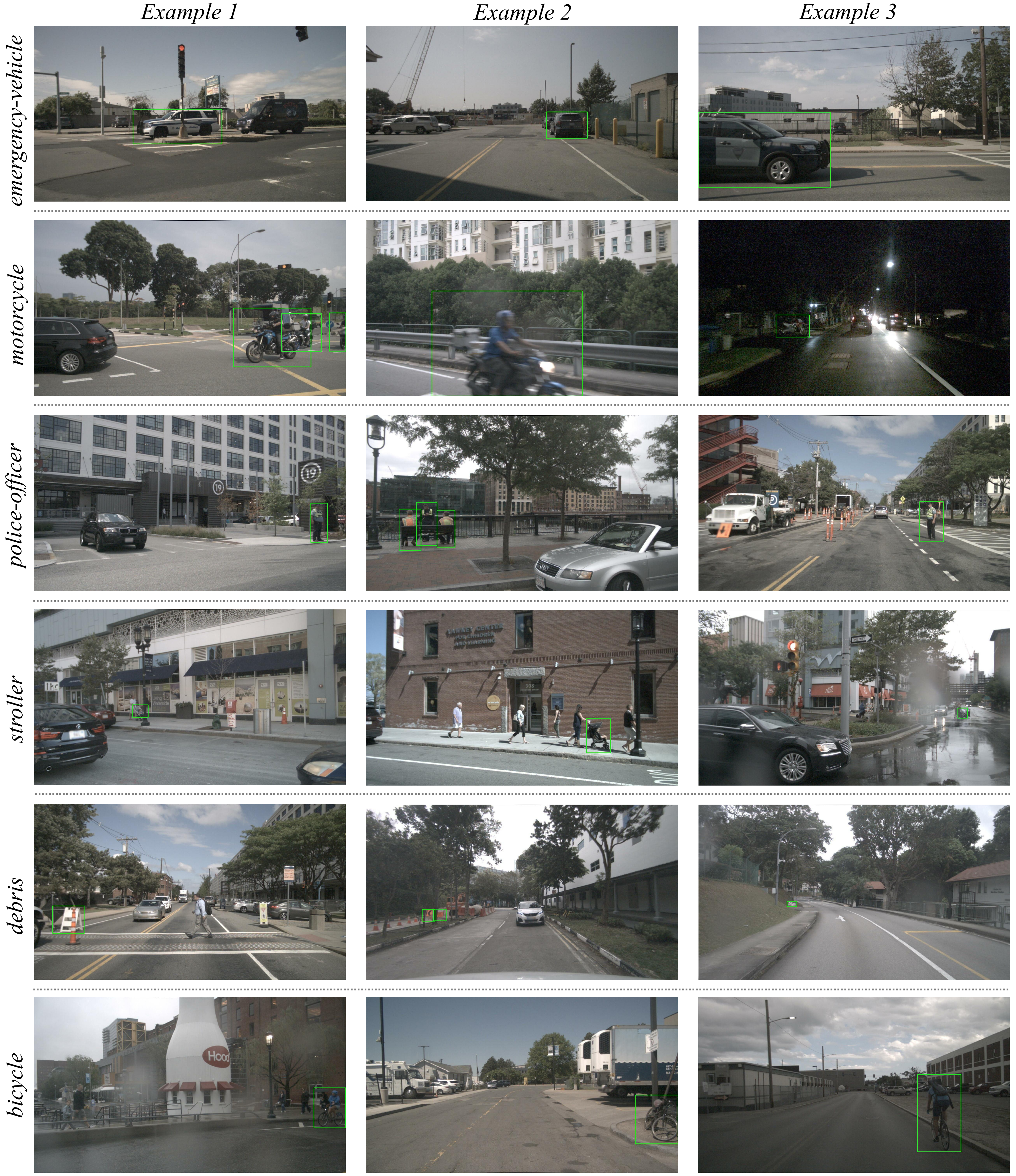}
\vspace{-2mm}
\caption{
\textbf{Visual demonstrations available in the annotation guidelines of the AutoExpert-nuScenes benchmark}.
We present 6 out of the 18 categories, with 3 example images shown for each category.
The green bounding boxes are 2D annotations for objects of the corresponding classes. 
It is worth noting the federated annotation: taking the {\tt emergency}-\texttt{vehicle} category as an example, even if objects of the {\tt car} category appear in the images, no corresponding annotations are provided.
}
\vspace{-4mm}
\label{fig:images_examples_guidelines}
\end{figure*}

\begin{figure*}[t]
\centering
\includegraphics[width=1.0\linewidth]{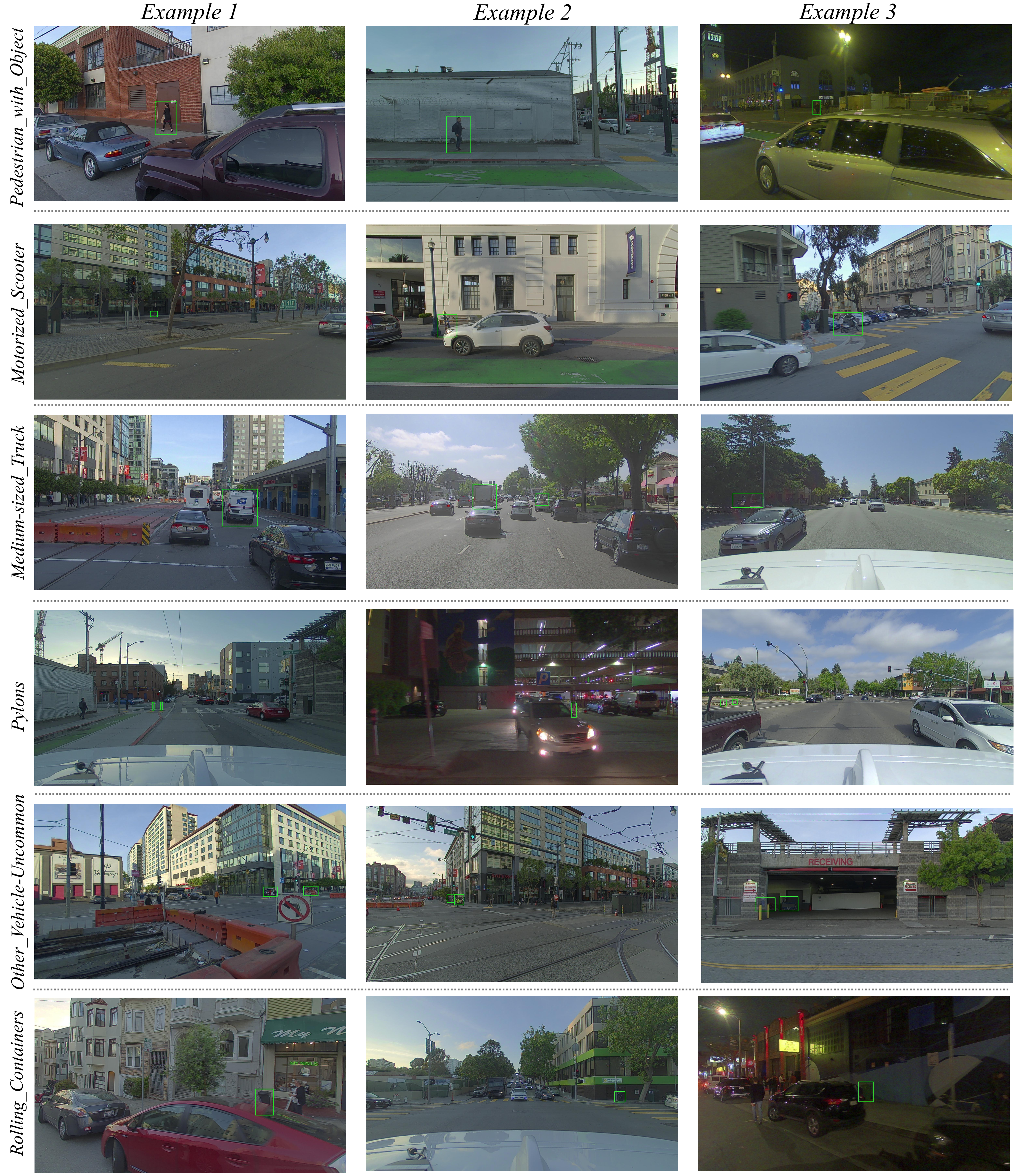}
\vspace{-2mm}
\caption{
\textbf{Visual demonstrations available in the annotation guidelines of the AutoExpert-PandaSet benchmark}.
We present 6 out of the 25 categories, with 3 example images shown for each category.
The green bounding boxes are 2D annotations for objects of the corresponding classes. 
It is worth noting the federated annotation: 
taking the \texttt{Medium-sized\_Truck} category as an example, even if objects of the \texttt{Car} or \texttt{Pedestrian} categories appear in the images, annotations do not contain them.
}
\vspace{-4mm}
\label{fig:images_examples_guidelines_pandaset}
\end{figure*}

Fig.~\ref{fig:images_examples_guidelines} and Fig.~\ref{fig:images_examples_guidelines_pandaset} illustrate representative image examples derived from the annotation guidelines of the AutoExpert-nuScenes and AutoExpert-PandaSet benchmarks, respectively. 
Specifically,
we visualize 6 out of 18 categories and 6 out of 25 categories for the two benchmarks, with 3 exemplar images shown for each category. 
A crucial characteristic of these training examples across both datasets is their \textit{federated annotation} style \cite{madan2024revisiting} --- only objects belonging to the specific target category are annotated with 2D bounding boxes, while other valid objects are intentionally ignored.
For instance, in the \texttt{emergency-vehicle} (of nuScenes) or \texttt{Medium-sized\_Truck} (of PandaSet) examples, adjacent cars or pedestrians remain unannotated. 
The complete set of these exemplar images for both benchmarks is provided in the supplementary material. 
Together with the expert-crafted descriptions, these images serve as the exclusive few-shot training data for the Multimodal Few-Shot Finetuning (MMFF) of our foundational 2D detector.

\section{Open-Source Code and Environments}
\label{sec:open-source-code-experimental-details-appendix}

\textbf{Open-source code and data.}
We include our codebase in the supplementary material and release it on \href{https://github.com/cc50121/AutoExpert-3D-benchmark}{GitHub}.
Refer to the \texttt{README.md} file for detailed instructions on setting up the environment and running the code.
In the supplementary material,
we also include the few-shot annotated images of the AutoExpert-nuScenes benchmark in the \texttt{nuScenes/} directory.
As the data size of 
AutoExpert-PandaSet training images is too large that exceeds supplement limit (100MB), we do not include these images in the supplement. 
Instead, we have released them in the 
\href{https://huggingface.co/datasets/autoexpert-cvpr2026-workshop/dataset-public}{HuggingFace repository}.



\textbf{Environments.}
Our development and evaluation environment is built upon Python 3.10.19 and PyTorch 2.9.0+cu118, utilizing 4 compute workers per GPU for efficient data loading. 
For optimizing the models, we adopt the AdamW optimizer with a learning rate of 1e-4 and a weight decay of 1e-4. 
To accommodate the computational demands of both the 2D detector training and the VLM reasoning, we utilize a compute node with four NVIDIA A100 GPUs. 
Specifically, for deploying the VLM, we integrate the vLLM framework (version 0.13.0) to achieve high-throughput and memory-efficient batched inference. 
Thanks to this multi-GPU data-parallel setup, the end-to-end finetuning of GroundingDINO (including the per-epoch validation) is highly accelerated,
taking approximately 15 minutes per epoch.

\newpage

\newpage
\section*{NeurIPS Paper Checklist}

\begin{enumerate}

\item {\bf Claims}
    \item[] Question: Do the main claims made in the abstract and introduction accurately reflect the paper's contributions and scope?
    \item[] Answer: \answerYes{} 
    \item[] Justification: Both the abstract and introduction clearly state our three major contributions, as itemized on Page 2 of this paper.
    \item[] Guidelines:
    \begin{itemize}
        \item The answer NA means that the abstract and introduction do not include the claims made in the paper.
        \item The abstract and/or introduction should clearly state the claims made, including the contributions made in the paper and important assumptions and limitations. A No or NA answer to this question will not be perceived well by the reviewers. 
        \item The claims made should match theoretical and experimental results, and reflect how much the results can be expected to generalize to other settings. 
        \item It is fine to include aspirational goals as motivation as long as it is clear that these goals are not attained by the paper. 
    \end{itemize}

\item {\bf Limitations}
    \item[] Question: Does the paper discuss the limitations of the work performed by the authors?
    \item[] Answer: \answerYes{} 
    \item[] Justification: \Cref{sec:discussions} discusses the limitations of this work.
    \item[] Guidelines:
    \begin{itemize}
        \item The answer NA means that the paper has no limitation while the answer No means that the paper has limitations, but those are not discussed in the paper. 
        \item The authors are encouraged to create a separate "Limitations" section in their paper.
        \item The paper should point out any strong assumptions and how robust the results are to violations of these assumptions (e.g., independence assumptions, noiseless settings, model well-specification, asymptotic approximations only holding locally). The authors should reflect on how these assumptions might be violated in practice and what the implications would be.
        \item The authors should reflect on the scope of the claims made, e.g., if the approach was only tested on a few datasets or with a few runs. In general, empirical results often depend on implicit assumptions, which should be articulated.
        \item The authors should reflect on the factors that influence the performance of the approach. For example, a facial recognition algorithm may perform poorly when image resolution is low or images are taken in low lighting. Or a speech-to-text system might not be used reliably to provide closed captions for online lectures because it fails to handle technical jargon.
        \item The authors should discuss the computational efficiency of the proposed algorithms and how they scale with dataset size.
        \item If applicable, the authors should discuss possible limitations of their approach to address problems of privacy and fairness.
        \item While the authors might fear that complete honesty about limitations might be used by reviewers as grounds for rejection, a worse outcome might be that reviewers discover limitations that aren't acknowledged in the paper. The authors should use their best judgment and recognize that individual actions in favor of transparency play an important role in developing norms that preserve the integrity of the community. Reviewers will be specifically instructed to not penalize honesty concerning limitations.
    \end{itemize}

\item {\bf Theory assumptions and proofs}
    \item[] Question: For each theoretical result, does the paper provide the full set of assumptions and a complete (and correct) proof?
    \item[] Answer: \answerNA{} 
    \item[] Justification: This paper does not contain theoretical results.
    \item[] Guidelines:
    \begin{itemize}
        \item The answer NA means that the paper does not include theoretical results. 
        \item All the theorems, formulas, and proofs in the paper should be numbered and cross-referenced.
        \item All assumptions should be clearly stated or referenced in the statement of any theorems.
        \item The proofs can either appear in the main paper or the supplemental material, but if they appear in the supplemental material, the authors are encouraged to provide a short proof sketch to provide intuition. 
        \item Inversely, any informal proof provided in the core of the paper should be complemented by formal proofs provided in appendix or supplemental material.
        \item Theorems and Lemmas that the proof relies upon should be properly referenced. 
    \end{itemize}

    \item {\bf Experimental result reproducibility}
    \item[] Question: Does the paper fully disclose all the information needed to reproduce the main experimental results of the paper to the extent that it affects the main claims and/or conclusions of the paper (regardless of whether the code and data are provided or not)?
    \item[] Answer: \answerYes{} 
    \item[] Justification: 
    With details described in \Cref{sec:problem-protocol} for benchmarking protocol and dataset preparation,
    we released our code and data in \href{https://github.com/cc50121/AutoExpert-3D-benchmark}{GitHub}, \href{https://huggingface.co/spaces/autoexpert-cvpr2026-workshop/auto3D}{HuggingFace}, respectively,
    as well as in the supplementary material,
    intending to guarantee reproducibility our our methods and fairly benchmark future methods.
    \item[] Guidelines:
    \begin{itemize}
        \item The answer NA means that the paper does not include experiments.
        \item If the paper includes experiments, a No answer to this question will not be perceived well by the reviewers: Making the paper reproducible is important, regardless of whether the code and data are provided or not.
        \item If the contribution is a dataset and/or model, the authors should describe the steps taken to make their results reproducible or verifiable. 
        \item Depending on the contribution, reproducibility can be accomplished in various ways. For example, if the contribution is a novel architecture, describing the architecture fully might suffice, or if the contribution is a specific model and empirical evaluation, it may be necessary to either make it possible for others to replicate the model with the same dataset, or provide access to the model. In general. releasing code and data is often one good way to accomplish this, but reproducibility can also be provided via detailed instructions for how to replicate the results, access to a hosted model (e.g., in the case of a large language model), releasing of a model checkpoint, or other means that are appropriate to the research performed.
        \item While NeurIPS does not require releasing code, the conference does require all submissions to provide some reasonable avenue for reproducibility, which may depend on the nature of the contribution. For example
        \begin{enumerate}
            \item If the contribution is primarily a new algorithm, the paper should make it clear how to reproduce that algorithm.
            \item If the contribution is primarily a new model architecture, the paper should describe the architecture clearly and fully.
            \item If the contribution is a new model (e.g., a large language model), then there should either be a way to access this model for reproducing the results or a way to reproduce the model (e.g., with an open-source dataset or instructions for how to construct the dataset).
            \item We recognize that reproducibility may be tricky in some cases, in which case authors are welcome to describe the particular way they provide for reproducibility. In the case of closed-source models, it may be that access to the model is limited in some way (e.g., to registered users), but it should be possible for other researchers to have some path to reproducing or verifying the results.
        \end{enumerate}
    \end{itemize}

\item {\bf Open access to data and code}
    \item[] Question: Does the paper provide open access to the data and code, with sufficient instructions to faithfully reproduce the main experimental results, as described in supplemental material?
    \item[] Answer: \answerYes{} 
    \item[] Justification: 
    We have released our code in \href{https://github.com/cc50121/AutoExpert-3D-benchmark}{GitHub} and the data in \href{https://huggingface.co/spaces/autoexpert-cvpr2026-workshop/auto3D}{HuggingFace}, respectively.
    \item[] Guidelines:
    \begin{itemize}
        \item The answer NA means that paper does not include experiments requiring code.
        \item Please see the NeurIPS code and data submission guidelines (\url{https://nips.cc/public/guides/CodeSubmissionPolicy}) for more details.
        \item While we encourage the release of code and data, we understand that this might not be possible, so ``No'' is an acceptable answer. Papers cannot be rejected simply for not including code, unless this is central to the contribution (e.g., for a new open-source benchmark).
        \item The instructions should contain the exact command and environment needed to run to reproduce the results. See the NeurIPS code and data submission guidelines (\url{https://nips.cc/public/guides/CodeSubmissionPolicy}) for more details.
        \item The authors should provide instructions on data access and preparation, including how to access the raw data, preprocessed data, intermediate data, and generated data, etc.
        \item The authors should provide scripts to reproduce all experimental results for the new proposed method and baselines. If only a subset of experiments are reproducible, they should state which ones are omitted from the script and why.
        \item At submission time, to preserve anonymity, the authors should release anonymized versions (if applicable).
        \item Providing as much information as possible in supplemental material (appended to the paper) is recommended, but including URLs to data and code is permitted.
    \end{itemize}

\item {\bf Experimental setting/details}
    \item[] Question: Does the paper specify all the training and test details (e.g., data splits, hyperparameters, how they were chosen, type of optimizer, etc.) necessary to understand the results?
    \item[] Answer: \answerYes{} 
    \item[] Justification: 
    \Cref{sec:problem-protocol} describes data splits for training, validation, and testing.
    The appendix provides more details on experiments.
    Importantly, our released code in \href{https://github.com/cc50121/AutoExpert-3D-benchmark}{GitHub} and data in \href{https://huggingface.co/spaces/autoexpert-cvpr2026-workshop/auto3D}{HuggingFace} contain all implementation details that warrant reproducibility.
    \item[] Guidelines:
    \begin{itemize}
        \item The answer NA means that the paper does not include experiments.
        \item The experimental setting should be presented in the core of the paper to a level of detail that is necessary to appreciate the results and make sense of them.
        \item The full details can be provided either with the code, in appendix, or as supplemental material.
    \end{itemize}

\item {\bf Experiment statistical significance}
    \item[] Question: Does the paper report error bars suitably and correctly defined or other appropriate information about the statistical significance of the experiments?
    \item[] Answer: \answerNo{} 
    \item[] Justification: 
    This paper prepares fixed data splits for training, validation and test sets.
    That said, it does not adopt random data splits.
    Moreover, as model training can start from random initializations,
    we tested different random seeds but observed little affects through 2D foundational detector finetuning. Therefore, the paper does not report error bars.
    \item[] Guidelines:
    \begin{itemize}
        \item The answer NA means that the paper does not include experiments.
        \item The authors should answer "Yes" if the results are accompanied by error bars, confidence intervals, or statistical significance tests, at least for the experiments that support the main claims of the paper.
        \item The factors of variability that the error bars are capturing should be clearly stated (for example, train/test split, initialization, random drawing of some parameter, or overall run with given experimental conditions).
        \item The method for calculating the error bars should be explained (closed form formula, call to a library function, bootstrap, etc.)
        \item The assumptions made should be given (e.g., Normally distributed errors).
        \item It should be clear whether the error bar is the standard deviation or the standard error of the mean.
        \item It is OK to report 1-sigma error bars, but one should state it. The authors should preferably report a 2-sigma error bar than state that they have a 96\% CI, if the hypothesis of Normality of errors is not verified.
        \item For asymmetric distributions, the authors should be careful not to show in tables or figures symmetric error bars that would yield results that are out of range (e.g. negative error rates).
        \item If error bars are reported in tables or plots, The authors should explain in the text how they were calculated and reference the corresponding figures or tables in the text.
    \end{itemize}

\item {\bf Experiments compute resources}
    \item[] Question: For each experiment, does the paper provide sufficient information on the computer resources (type of compute workers, memory, time of execution) needed to reproduce the experiments?
    \item[] Answer: \answerYes{} 
    \item[] Justification: \Cref{sec:exp} and Appendix \S\ref{sec:open-source-code-experimental-details-appendix} clearly state computing resources, e.g., using a computing node with four NVIDIA A100 GPUs.
    Appendix \S\ref{sec:comp_efficiency} further analyzes time costs.
    \item[] Guidelines:
    \begin{itemize}
        \item The answer NA means that the paper does not include experiments.
        \item The paper should indicate the type of compute workers CPU or GPU, internal cluster, or cloud provider, including relevant memory and storage.
        \item The paper should provide the amount of compute required for each of the individual experimental runs as well as estimate the total compute. 
        \item The paper should disclose whether the full research project required more compute than the experiments reported in the paper (e.g., preliminary or failed experiments that didn't make it into the paper). 
    \end{itemize}
    
\item {\bf Code of ethics}
    \item[] Question: Does the research conducted in the paper conform, in every respect, with the NeurIPS Code of Ethics \url{https://neurips.cc/public/EthicsGuidelines}?
    \item[] Answer: \answerYes{} 
    \item[] Justification: We read the guidelines careful and confirm that the research in this paper conform with the NeurIPS Code of Ethics.
    \item[] Guidelines:
    \begin{itemize}
        \item The answer NA means that the authors have not reviewed the NeurIPS Code of Ethics.
        \item If the authors answer No, they should explain the special circumstances that require a deviation from the Code of Ethics.
        \item The authors should make sure to preserve anonymity (e.g., if there is a special consideration due to laws or regulations in their jurisdiction).
    \end{itemize}

\item {\bf Broader impacts}
    \item[] Question: Does the paper discuss both potential positive societal impacts and negative societal impacts of the work performed?
    \item[] Answer: \answerYes{} 
    \item[] Justification: \Cref{sec:discussions} discusses broader impacts including both potential positive societal impacts and negative societal impacts.
    \item[] Guidelines:
    \begin{itemize}
        \item The answer NA means that there is no societal impact of the work performed.
        \item If the authors answer NA or No, they should explain why their work has no societal impact or why the paper does not address societal impact.
        \item Examples of negative societal impacts include potential malicious or unintended uses (e.g., disinformation, generating fake profiles, surveillance), fairness considerations (e.g., deployment of technologies that could make decisions that unfairly impact specific groups), privacy considerations, and security considerations.
        \item The conference expects that many papers will be foundational research and not tied to particular applications, let alone deployments. However, if there is a direct path to any negative applications, the authors should point it out. For example, it is legitimate to point out that an improvement in the quality of generative models could be used to generate deepfakes for disinformation. On the other hand, it is not needed to point out that a generic algorithm for optimizing neural networks could enable people to train models that generate Deepfakes faster.
        \item The authors should consider possible harms that could arise when the technology is being used as intended and functioning correctly, harms that could arise when the technology is being used as intended but gives incorrect results, and harms following from (intentional or unintentional) misuse of the technology.
        \item If there are negative societal impacts, the authors could also discuss possible mitigation strategies (e.g., gated release of models, providing defenses in addition to attacks, mechanisms for monitoring misuse, mechanisms to monitor how a system learns from feedback over time, improving the efficiency and accessibility of ML).
    \end{itemize}
    
\item {\bf Safeguards}
    \item[] Question: Does the paper describe safeguards that have been put in place for responsible release of data or models that have a high risk for misuse (e.g., pretrained language models, image generators, or scraped datasets)?
    \item[] Answer: \answerNA{} 
    \item[] Justification: The data used in this paper are from established autonomous driving datasets that do not have a high risk for misuse.
    \item[] Guidelines:
    \begin{itemize}
        \item The answer NA means that the paper poses no such risks.
        \item Released models that have a high risk for misuse or dual-use should be released with necessary safeguards to allow for controlled use of the model, for example by requiring that users adhere to usage guidelines or restrictions to access the model or implementing safety filters. 
        \item Datasets that have been scraped from the Internet could pose safety risks. The authors should describe how they avoided releasing unsafe images.
        \item We recognize that providing effective safeguards is challenging, and many papers do not require this, but we encourage authors to take this into account and make a best faith effort.
    \end{itemize}

\item {\bf Licenses for existing assets}
    \item[] Question: Are the creators or original owners of assets (e.g., code, data, models), used in the paper, properly credited and are the license and terms of use explicitly mentioned and properly respected?
    \item[] Answer: \answerYes{} 
    \item[] Justification: \Cref{sec:problem-protocol} clearly states that we repurpose the established nuScenes and PandaSet datasets, which are is publicly available under  the CC BY-NC-SA 4.0 license.
    Our code and annotations are licensed under the MIT License.
    \item[] Guidelines:
    \begin{itemize}
        \item The answer NA means that the paper does not use existing assets.
        \item The authors should cite the original paper that produced the code package or dataset.
        \item The authors should state which version of the asset is used and, if possible, include a URL.
        \item The name of the license (e.g., CC-BY 4.0) should be included for each asset.
        \item For scraped data from a particular source (e.g., website), the copyright and terms of service of that source should be provided.
        \item If assets are released, the license, copyright information, and terms of use in the package should be provided. For popular datasets, \url{paperswithcode.com/datasets} has curated licenses for some datasets. Their licensing guide can help determine the license of a dataset.
        \item For existing datasets that are re-packaged, both the original license and the license of the derived asset (if it has changed) should be provided.
        \item If this information is not available online, the authors are encouraged to reach out to the asset's creators.
    \end{itemize}

\item {\bf New assets}
    \item[] Question: Are new assets introduced in the paper well documented and is the documentation provided alongside the assets?
    \item[] Answer: \answerYes{} 
    \item[] Justification: 
    The new assets in this paper contain our code, prepared data,
    and annotations.
    \Cref{sec:problem-protocol} and Appendix \S\ref{sec:pandaset-appendix} describe the data in detail.
    We have released our code in \href{https://github.com/cc50121/AutoExpert-3D-benchmark}{GitHub} and data in \href{https://huggingface.co/spaces/autoexpert-cvpr2026-workshop/auto3D}{HuggingFace}.
    \item[] Guidelines:
    \begin{itemize}
        \item The answer NA means that the paper does not release new assets.
        \item Researchers should communicate the details of the dataset/code/model as part of their submissions via structured templates. This includes details about training, license, limitations, etc. 
        \item The paper should discuss whether and how consent was obtained from people whose asset is used.
        \item At submission time, remember to anonymize your assets (if applicable). You can either create an anonymized URL or include an anonymized zip file.
    \end{itemize}

\item {\bf Crowdsourcing and research with human subjects}
    \item[] Question: For crowdsourcing experiments and research with human subjects, does the paper include the full text of instructions given to participants and screenshots, if applicable, as well as details about compensation (if any)? 
    \item[] Answer: \answerNA{} 
    \item[] Justification: The paper does not involve crowdsourcing nor research with human subjects.
    \item[] Guidelines:
    \begin{itemize}
        \item The answer NA means that the paper does not involve crowdsourcing nor research with human subjects.
        \item Including this information in the supplemental material is fine, but if the main contribution of the paper involves human subjects, then as much detail as possible should be included in the main paper. 
        \item According to the NeurIPS Code of Ethics, workers involved in data collection, curation, or other labor should be paid at least the minimum wage in the country of the data collector. 
    \end{itemize}

\item {\bf Institutional review board (IRB) approvals or equivalent for research with human subjects}
    \item[] Question: Does the paper describe potential risks incurred by study participants, whether such risks were disclosed to the subjects, and whether Institutional Review Board (IRB) approvals (or an equivalent approval/review based on the requirements of your country or institution) were obtained?
    \item[] Answer: \answerNA{} 
    \item[] Justification: The paper does not require IRB approval as it does not study  human subjects.
    \item[] Guidelines:
    \begin{itemize}
        \item The answer NA means that the paper does not involve crowdsourcing nor research with human subjects.
        \item Depending on the country in which research is conducted, IRB approval (or equivalent) may be required for any human subjects research. If you obtained IRB approval, you should clearly state this in the paper. 
        \item We recognize that the procedures for this may vary significantly between institutions and locations, and we expect authors to adhere to the NeurIPS Code of Ethics and the guidelines for their institution. 
        \item For initial submissions, do not include any information that would break anonymity (if applicable), such as the institution conducting the review.
    \end{itemize}

\item {\bf Declaration of LLM usage}
    \item[] Question: Does the paper describe the usage of LLMs if it is an important, original, or non-standard component of the core methods in this research? Note that if the LLM is used only for writing, editing, or formatting purposes and does not impact the core methodology, scientific rigorousness, or originality of the research, declaration is not required.
    \item[] Answer: \answerYes{} 
    \item[] Justification:
    We use LLMs to polish writing.
    Other than LLMs,
    we also use Vision-Language Models (VLMs) and Foundational Vision Models (FVMs) as technical components in our developed methods.
    \Cref{sec:method} clearly states how they are used in the developed methods. 
    That said, we did not use LLMs and other AI agents to impact the core methodology, scientific rigorousness or originality of our research.
    \item[] Guidelines:
    \begin{itemize}
        \item The answer NA means that the core method development in this research does not involve LLMs as any important, original, or non-standard components.
        \item Please refer to our LLM policy (\url{https://neurips.cc/Conferences/2025/LLM}) for what should or should not be described.
    \end{itemize}

\end{enumerate}

\end{document}